\documentclass{article}

\usepackage{PRIMEarxiv}

\usepackage[utf8]{inputenc} 
\usepackage[T1]{fontenc}    
\usepackage{hyperref}       
\usepackage{url}            
\usepackage{booktabs}       
\usepackage{amsfonts}       
\usepackage{nicefrac}       
\usepackage{microtype}      
\usepackage{lipsum}
\usepackage{fancyhdr}       
\usepackage{graphicx}       

\usepackage{amsmath,amssymb,amsfonts}
\usepackage{algorithmic}
\usepackage{graphicx}
\usepackage{textcomp}
\usepackage{xcolor}

\usepackage[ruled,vlined,linesnumbered]{algorithm2e}
\usepackage{booktabs}
\usepackage{tabularx}
\usepackage{multirow}
\usepackage{url}
\usepackage[multiple]{footmisc}

\newcommand{\RQA}{How to design an automated and generic hybrid forecasting approach that combines different forecasting methods to compensate for the disadvantages of each technique?}

\newcommand{\RQB}{How to automatically extract and transform features of the considered
time series to increase the forecast accuracy?}

\newcommand{\RQC}{What are appropriate strategies to dynamically apply the most accurate
method within the hybrid forecasting approach for a given time series?}

\title{Telescope: An Automated Hybrid Forecasting Approach on a Level-Playing Field
}

\author{
  André Bauer \\
  University of Chicago \\
  Chicago, United States of America\\
  \texttt{andrebauer@uchicago.edu} \\
  \And
  Mark Leznik \\
  University of Ulm \\
  Ulm, Germany\\
  \texttt{mark.leznik@uni-ulm.de} \\
   \And
  Michael Stenger, Robert Leppich, Nikolas Herbst, Samuel Kounev \\
  Affiliation \\
  City\\
  \texttt{\{firstname\}.\{lastname\}@uni-wuerzburg.de} \\
  \And
  Ian Foster \\
  University of Chicago \\
  Chicago, United States of America\\
  \texttt{foster@cs.uchicago.edu}
}

\begin{document}
\maketitle

\begin{abstract}
In many areas of decision-making, forecasting is an essential pillar. Consequently, many different forecasting methods have been proposed. 
From our experience, recently presented forecasting methods are computationally intensive, poorly automated, tailored to a particular data set, or they lack a predictable time-to-result. 
To this end, we introduce Telescope, a novel machine learning-based forecasting approach that automatically retrieves relevant information from a given time series and splits it into parts, handling each of them separately. In contrast to deep learning methods, our approach doesn't require parameterization or the need to train and fit a multitude of parameters. It operates with just one time series and provides forecasts within seconds without any additional setup.
Our experiments show that Telescope outperforms recent methods by providing accurate and reliable forecasts while making no assumptions about the analyzed time series.
\end{abstract}

\keywords{Automatic feature extraction \and Combining forecasts \and Comparative studies \and Forecasting competitions \and Long term time series forecasting \and Time series}

\section{Introduction}

Time series forecasting is an essential pillar in many decision-making disciplines~\cite{HA14}. Accordingly, time series forecasting is an established and active field of research, and thus various methods have been proposed. Due to the variety of approaches, the choice and configuration of the best performing method for a given time series remain a mandatory expert task to avoid trial-and-error. However, expert knowledge can be expensive, may have a subjected bias, and it can take a long time to deliver results. 

Thus, the question arises if there is a single forecasting method that performs best for all time series. However, the ``No-Free-Lunch Theorem''~\cite{NoFreeLunch}, initially formulated for optimization problems, denies the possibility of such a method. It states that improving one aspect typically leads to a degradation in performance for another aspect. An analogy can be drawn to the domain of forecasting: No forecasting method performs best for all time series. In other words, forecasting methods have their advantages and drawbacks depending on the considered time series.  

In fact, different types of hybrid forecasting methods have been proposed in the last years~\cite{hybridforecasts} to tackle the challenge stated by the ``No-Free-Lunch Theorem''. The core idea is the usage of at least two methods to minimize the disadvantages of individual methods. From our experience, recently presented open-source hybrid methods are computationally intensive, poorly automated, tailored to a particular data set, or they lack a predictable time-to-result. However, many real-world scenarios where forecasting is useful (e.g., auto-scaling) have strict requirements for a reliable time-to-result and forecast accuracy. To achieve a low variance in forecast accuracy, the preprocessing of historical data and the feature handling (extraction, engineering, and selection) must be done in a sophisticated way. 

To tackle the mentioned challenges, we pose ourselves the following research questions: 

RQ1~\textit{\RQA{}}

RQ2~\textit{\RQB{}}

RQ3~\textit{\RQC{}}

Towards addressing the research questions, we introduce a novel machine learning-based hybrid forecasting approach called \emph{Telescope}\footnote{Telescope at GitHub: \url{https://github.com/DescartesResearch/telescope}}\footnote{A preliminary idea of our Telescope approach were accepted as a short conference publication~\cite{BaZuHeKoCu-ICDE-Telescope}.} that automatically retrieves relevant information from a given time series and splits it into parts, handling each of them separately (addressing RQ1).  More precisely, Telescope automatically extracts intrinsic time series features and then decomposes the time series into components, building a forecasting model for each of them (addressing RQ2). Each component is forecast by applying a different method and then the final forecast is assembled from the forecast components by employing a regression-based machine learning algorithm. For non-time-critical scenarios, we additionally provide an internal recommendation system that can be employed to automatically select the most appropriate machine learning algorithm for assembling the time series from its components (addressing RQ3).

To evaluate our approach, we compare Telescope to seven competing methods (including approaches from Uber and Facebook) in more than 1000 hours of experiments using a forecasting benchmark. Telescope outperformed all methods, exhibiting the best forecast accuracy coupled with a low and reliable time-to-result. Compared to the competing methods that exhibited, on average, a forecast error (more precisely, the symmetric mean absolute error) of 29\%, Telescope exhibited an error of 20\% while being 2556~times faster. In particular, the methods from Uber and Facebook exhibited an error of 48\% and 36\%, and were 7334 and 19 times slower than Telescope, respectively.  When additionally applying the recommendation system, Telescope was able to reduce the forecast error even further down to 19\%. 

The rest of the article is structured as follows: We first introduce terms as well as definitions for understanding this article in Section~\ref{sec:foundations}. Then, we review related work in Section~\ref{sec:rw}. In Section~\ref{sec:approach}, we present Telescope. Afterward, we benchmark Telescope in Section~\ref{sec:results} before concluding this article.

\section{Foundations on Time Series}
\label{sec:foundations}

In this section, we briefly present the basic terms and definitions related to time series and time series in Section~\ref{foundations:sec:definitions}. Then, we outline the frequency detection of a time series, the time series decomposition, and the time series transformation in Section~\ref{foundations:sec:spectralanalysis}, \ref{foundations:sec:tsdecomposition}, and \ref{foundations:sec:tstransformation}, respectively.

\subsection{Terms and Definitions}
\label{foundations:sec:definitions}
A \emph{univariate time series} is an ordered collection of values of a quantity obtained over a specific period or since a certain point in time. In general, observations are recorded in successive and equidistant time steps (e.g., hours). Typically, internal patterns exist, such as autocorrelation, trend, or seasonal variation. Mathematically, if $y_t \in \mathbb{R}$ is the observed value at time $t$ and $T$ a discrete set of equidistant time points, a univariate time series is defined by 
\begin{align}
Y := \left\{y_t: t \in T\right\}.    
\end{align}

\subsubsection{Components of a Time Series}
\label{foundations:sec:components}

A time series can also be seen as a composition of trend, seasonal, cycle, and irregular components~\cite{HA14}. The long-term development in a time series (i.e., upwards, downwards, or stagnate) is called \emph{trend}. Usually, the trend is a monotonic function unless external events trigger a break and cause a change in the direction. The presence of recurring patterns within a regular period in the time series is called \emph{seasonality}. These patterns are typically caused by climate, customs, or traditional habits such as night and day phases. The length of a seasonal pattern is called frequency\footnote{In the context of time series analysis, the term frequency has a different meaning as, for example, in physics.}. Rises and falls within a time series without a fixed frequency are called \emph{cycles}. In contrast to seasonality, the amplitude and duration of the cycles vary over time. The remaining part of the time series that is not described by trend, seasonality, or cycles is called the \emph{irregular component}. It usually follows a certain statistical noise distribution and is therefore not predictable by most models. 

\subsubsection{Fourier Terms}
\label{foundations:sec:fourierterms}

Following the basic principle in mathematics to break down complex objects into more simpler parts, a time series can be approximated by a linear combination of sinusoid terms. The resulting representation is referred to as \emph{Fourier series} and the sinusoids terms are called \emph{Fourier terms}. 

\subsubsection{Stationarity}
\label{foundations:sec:stationarity}
One of the most important characteristics of a time series is the \emph{stationarity} since most forecasting methods assume that the time series is either stationary or can be ``stationarized''  through a transformation~\cite{brockwell2016introduction}. Loosely speaking, the statistical properties (such as mean, variance, and auto-covariance) of a stationary time series do not change over time. In practice, however, time series are usually showing a mix of trend or/and seasonal patterns and are thus non-stationary~\cite{adhikari2013introductory}. To this end, time series are transformed, seasonally adjusted, made trend-stationary by removing the trend, or made difference-stationary by possibly repeated differencing (i.e., computing the differences between consecutive observations)~\cite{HA14}.

\subsection{Spectral Analysis and Frequency Detection}
\label{foundations:sec:spectralanalysis}

In many fields, especially for forecasting, it helps discover the underlying periodicities in the data, that is, knowing the frequencies or the lengths of the seasonal patterns. For instance, if the most dominant frequency is unknown for a given time series, the time series cannot be decomposed (see Section~\ref{foundations:sec:tsdecomposition}). In this context, the dominant frequency means the most common period, respectively, the seasonal pattern, such as days in a year. Also, if the dominant frequency is available, the information on the next dominant frequency (e.g., the week within the year) is helpful. A standard method for this data analysis is the \emph{spectral analysis}, also referred to as analysis in the frequency domain. The key idea is to transform the time series from the time domain to the frequency domain as the spectrum reveals the data's underlying frequencies. The resulting \emph{spectral density} assigns then an intensity to each frequency within the time series. An estimate of the spectral density is the \emph{periodogram}~\cite{schuster1899periodgram}. That is, in a time series that is driven by certain seasonal patterns, the periodogram shows peaks at precisely those frequencies.

\subsection{Time Series Decomposition}
\label{foundations:sec:tsdecomposition}
As a time series consists of different components, a common approach is to break down the time series into its components. The parts can either be used to modify the data (e.g., removing the trend or seasonality) or be used as intrinsic features for augmenting a model capturing the time series. 

\begin{figure}[htb!]
	\centering
	\includegraphics[width=0.75\columnwidth]{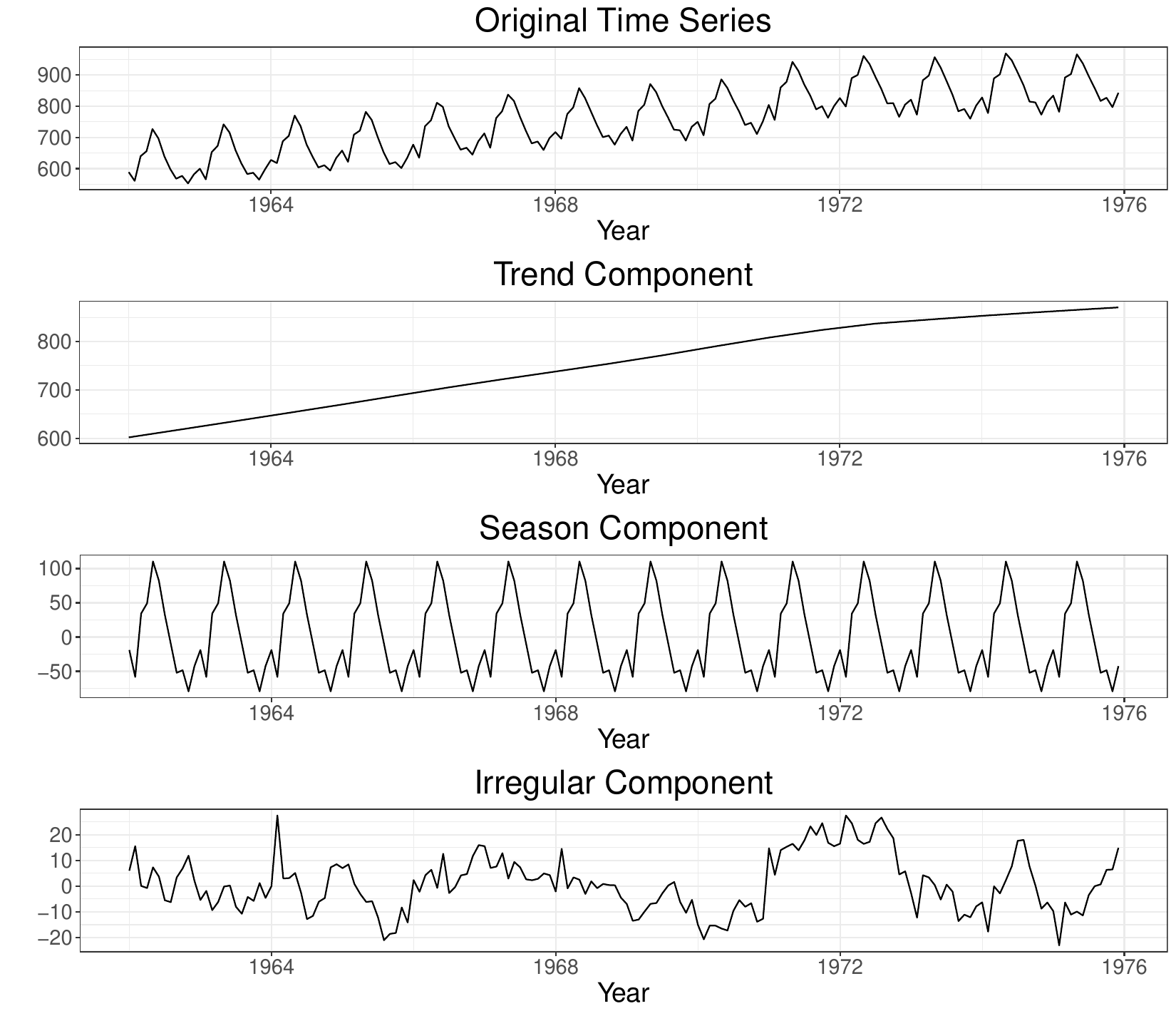}
	\caption{Example of STL decomposition.}
	\label{fig:decomp}
\end{figure}

A common method for decomposing a time series is \emph{STL} (\emph{S}easonal and \emph{T}rend decomposition using \emph{L}oess)~\cite{cleveland1990stl}. STL disassembles the given time series $Y(t)$ into the components trend $T(t)$, season $S(t)$, and irregular $I(t)$. More formally, the resulting decomposition of STL is 
\begin{align}
Y(t) := T(t) + S(t) + I(t). 
\end{align}
Although STL can only handle additive relationships between the components, it can also be applied to multiplicative time series by applying the natural logarithm on the time series beforehand. Figure \ref{fig:decomp} depicts a decomposition of a time series based on STL. 

\subsection{Time Series Transformation}
\label{foundations:sec:tstransformation}

As observed data may be quite complex, for example, having high variance and/or multiplicative relationship between the components, an adjustment or simplification of it can improve the forecasting model~\cite{HA14}. To this end, there are different methods that transform time series. A common approach is to apply the logarithm, but the transformed data may not be normally distributed. In contrast, the \emph{Box-Cox transformation}~\cite{box1964analysis} tries to transform the data into ``normal shape''. The Box-Cox transformation is defined as follows
\begin{align}
w_t  := \begin{cases} \ln y_t & \text{if $\lambda=0$,}  \\ (y_t^\lambda-1)/\lambda & \text{otherwise,} \end{cases} 
\end{align}
where $y_t$ is the original time series and $\lambda$ the transformation parameter that determines the function.

Note that if a forecast was conducted based on this transformation, the forecast values have to be re-transformed using the same $\lambda$ to be in the right scale. Consequently, the re-transformation of the time series is defined as 
\begin{align}
y_{t} := \begin{cases} e^{w_{t}} & \lambda=0 ,\\ (\lambda w_t+1)^{1/\lambda} & \text{otherwise.}\end{cases} 
\end{align}

\section{Related Work}
\label{sec:rw}

In 1997, the ``No-Free-Lunch Theorem'' was postulated. It states that there is not a single optimization algorithm that performs best for all scenarios since improving the performance of one aspect leads typically to a degradation in performance for another aspect. Considering the inherent drawbacks and limitations of forecasting methods, it can be concluded that there is no single forecasting method that performs best for all kinds of time series. To face this challenge, many hybrid forecasting methods have been proposed in the literature. The underlying idea of such hybrid approaches is to use at least two forecasting techniques to compensate for the limitations of the individual forecasting approaches. The success of this concept can be demonstrated, for example, by investigating the recent M4-Competition~\cite{makridakis2018m4}: 12 of the 17 most accurate methods were hybrid forecasting methods. The proposed hybrid methods can be categorized into three groups, each sharing the same basic concept: (i)~\emph{ensemble forecasting}, (ii)~\emph{forecasting method recommendation}, and (iii)~\emph{time series decomposition}. Consequently, Sections \ref{pt:foundations:ch:rwforecasting:sec:ensemble} to \ref{pt:foundations:ch:rwforecasting:sec:decomposition} present top cited and recent hybrid approaches. The delimitation of Telescope from the following approaches is discussed in Section~\ref{pt:contribution:ch:telescope:sec:delimitation}.

\subsection{Ensemble Forecasting} 
\label{pt:foundations:ch:rwforecasting:sec:ensemble}
The core idea of the first and historically oldest group is to compute the forecast as a weighted sum of the values derived from applying multiple forecasting methods. 
The impetus for this idea was provided by J.~Bates and C.~Granger in 1969~\cite{bates1969combination}. In their work, they show that the forecast accuracy can be increased by using ensembles compared to applying each forecasting method on its own. However, the accuracy of ensemble forecasting depends highly on the weight estimation algorithm. Recent work applies a weighted sum based on the training error~\cite{adhikari2015model} or starts with a simple arithmetic mean of the forecasts and learns then the optimal weights for each forecasting method dynamically~\cite{sommer2016local}. Another approach~\cite{cerqueira2017arbitrated} applies a meta-learner to determine the weights based on each method's forecast error based on the training set. Z.~Wang et~al.~\cite{wang2018static} propose an ensemble of 30 feed-forward neural networks, where each network gets a random subset of the time series. D.~Boulegane et al.~\cite{boulegane2019arbitrated} builds an ensemble base on Bernoulli trial that takes the predicted error and the meta-learner's confidence for each method into account. P.~Montero-Manso et~al.~\cite{montero2020fforma} applies a meta-learner that learns the ensemble weigths based time series characteristics and the forecast error of each method.

\subsection{Forecasting Method Recommendation} 
\label{pt:foundations:ch:rwforecasting:sec:recommendation}

Methods from the second group build a rule set for estimating the assumed best forecasting method based on analyzing specific characteristics of the considered time series characteristics. 
The first rule for weighting forecasting methods based on the characteristics of the given time series was introduced F.~Collopy and J.~Armstrong in 1992~\cite{collopy1992rule}. In their work, the authors manually created an expert system after having interviewed five experts in the field of forecasting from industry and academia. A few years later, X.~Wang et~al. used clustering and self-organizing maps to select the best forecasting method based on time series characteristics~\cite{Wang20092581}. This work was extend by A.~Widodo and I.~Budi~\cite{Widodo2013ModelSU} by adding more methods. In addition to time series characteristics, M.~K{\"u}ck et~al.\cite{kuck2016meta} considered for selecting the best forecasting method for a given time series a set of error measures. In another approach~\cite{lemke2010meta2}, the time series are classified into different clusters and the method that has the lowest prediction error in the cluster in which the new time series is classified is recommended. T.~Talagala et~al.~\cite{talagala2018meta} applied a meta-learner to map the best forecasting method to a given time series based on its characteristics. A similar approach is proposed by D.~Zhang et al.~\cite{zhang2020forecasting}, where the forecast horizon is also considered for the meta-learning.

\subsection{Time Series Decomposition} 
\label{pt:foundations:ch:rwforecasting:sec:decomposition}

In the last group, a time series is decomposed into components and forecasting methods are applied to each component separately or a time series is forecast with an individual method, and afterward, a second individual method is applied on the residuals. 
Recent approaches decompose the time series into a linear and non-linear part based on fitting residuals~\cite{zhang2003time, panigrahi2017hybrid}, discrete wavelet transformation~\cite{khandelwal2015time}, or empirical mode decomposition~\cite{liu2014hybrid,zhang2017hybrid}. C.~Bergmeir et~al.~\cite{bergmeir2016bagging} decomposed the time series with STL and then, bootstrapped different versions of the irregular part to form new time series. Each new time series is forecast and the final forecast is the median of these forecasts. S.~Taylor and B.~Letham~(Facebook)~\cite{taylor2018forecasting} split the time series into trend, season, and holiday. F. Sa{\^a}daoui and H. Rabbouch proposed an approach that splits a time series into trend, seasonal, and irregular parts based on the maximal overlap discrete wavelet transformation~\cite{saadaoui2019wavelet} and one approach approximating the trend with a linear regression and seasonal component with a sum of sines~\cite{saadaoui2019hybrid}. The method~\cite{smyl2020hybrid} introduced by S.~Syml~(Uber) deploys exponential smoothing for de-seasonalizing and a a recurrent neural network for extrapolating the time series.

\subsection{Differentiation from Related Work}
\label{pt:contribution:ch:telescope:sec:delimitation}

One of the key differences to the proposed hybrid methods~\cite{wang2018static,collopy1992rule,zhang2020forecasting,liu2014hybrid,bergmeir2016bagging,taylor2018forecasting,saadaoui2019wavelet,smyl2020hybrid} is that Telescope is generic. That is, Telescope is not tailored to a special scenario and makes no assumptions about the analyzed time series. Further, as Telescope is based on time series decomposition and its recommendation part is only used in non-time-critical scenarios, our approach differs considerably from methods from the field of ensemble forecasting. In contrast to the methods originating from time series decomposition that use different decomposition and forecasting techniques, Telescope explicitly decomposes the time series into trend, season, and irregular part. Also, each part is forecast separately using different forecasting methods. In contrast to the recommendation-based methods that use mainly ``classical'' time series forecasting methods, Telescope selects regression-based machine learning methods based on time series characteristics. Lastly, our approach augments the original time series set by generating new time series from it to increase the data set's diversity.

\section{The Telescope Approach}
\label{sec:approach}

The assumption of data stationarity is an inherent limitation for time series forecasting. Any time series property that eludes stationarity, such as non-constant mean (i.e., trend), seasonality, non-constant variance, or multiplicative effect, poses a challenge for the proper model building~\cite{makridakis2018statistical}. Consequently, we take all the techniques discussed in the previous section into account to design an automated forecasting workflow called Telescope that automatically transforms the given time series, derives intrinsic features from the time series, selects a suitable set of features, and handles each feature separately. The choice and combinations of different methods and techniques contains the best-performing methods during preliminary experiments.

\begin{algorithm}[htb!]
	\small
	\newcommand\mycommfont[1]{\footnotesize\ttfamily\textcolor{darkgray}{#1}}
	\SetCommentSty{mycommfont}
	
	\caption{Telescope forecasting workflow.}
	\label{algo:telescope:overview} 
	\SetAlgoLined
	
	\KwIn{Time series \textit{ts}, horizon \textit{h}}
	\KwResult{Forecast of \textit{ts}}
	[\textit{ts}, \textit{freqs}] = Preprocessing(\textit{ts})\tcp*[l]{see Section~\ref{approach:sec:preprocessing}} 
	\uIf(\tcp*[h]{ts is seasonal}){\textit{freqs}[1] $>$ 1}{
		\textit{features} = FeatureExtraction(\textit{ts})\tcp*[l]{see Section~\ref{approach:sec:featureextraction}}
		\textit{model} = ModelBuilding(\textit{ts}, \textit{features})\tcp*[l]{see Section~\ref{approach:sec:modelbuilding}}
		\textit{forecast} = Forecasting(\textit{model}, \textit{h})\tcp*[l]{see Section~\ref{approach:sec:forecasting}}
	}
	\Else{
		\textit{forecast} = ARIMA(\textit{ts}, \textit{h})\tcp*[l]{see Section~\ref{approach:sec:fallback}}
	}
	\textit{forecast} = Postprocessing(\textit{forecast})\tcp*[l]{see Section~\ref{approach:sec:postprocessing}}
	
	\KwRet{\textit{forecast}}
	
\end{algorithm}

The workflow of Telescope is briefly illustrated in Algorithm~\ref{algo:telescope:overview} and gets as input a univariate time series \textit{ts} and the horizon \textit{h}. The horizon specifies how many values have to be forecast at once. In the first phase (Line~1), the time series is preprocessed and the frequencies of the underlying patterns are extracted. Telescope is intended to handle seasonal time series as many time series are observed or produced by systems subjected to human habits and are thus seasonal. In other words, if a seasonal time series has to be forecast, the second and third phases of Telescope comprise the extracting of relevant intrinsic time series features (Line~3) and building a model that describes the time series based on these features (Line~4). Afterward, the model is used to forecast the behavior of the future time series (Line~5). In the case where no seasonality exists within a time series (Line 7), the time series is modeled and forecast with ARIMA~\cite{HA14}. Finally, the forecast is postprocessed according to the preprocessing phase and returned. In the following, each phase is explained in detail. Afterwards, the limitations and assumptions are discussed in Section~\ref{approach:sec:limitations}.

\subsection{Preprocessing}
\label{approach:sec:preprocessing}

\begin{figure}[htb!]
	\centering
	\includegraphics[width=\linewidth]{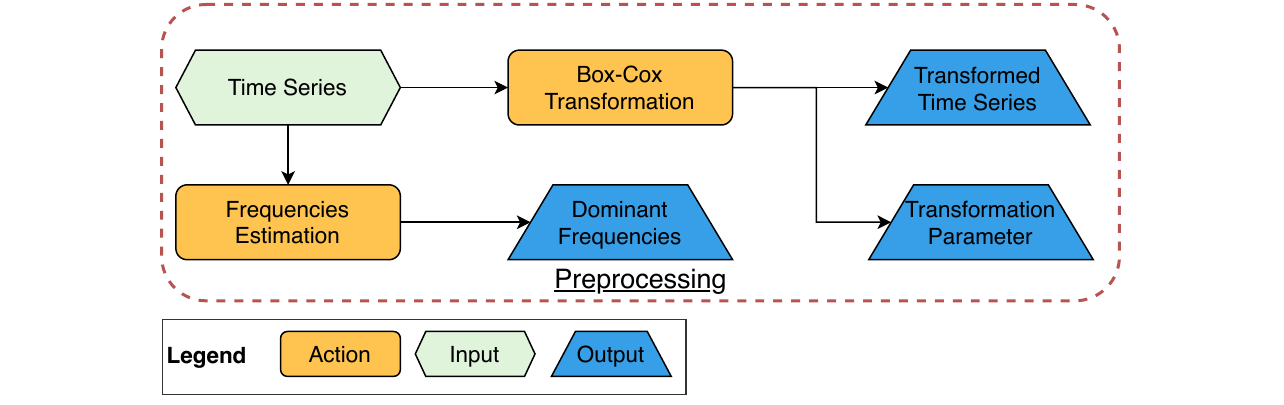}
	\caption{Preprocessing phase of Telescope.}
	\label{fig:tel:preprocessing}
\end{figure}

The first phase of Telescope is called \textit{Preprocessing} and the workflow is depicted in Figure~\ref{fig:tel:preprocessing}. Orange, rounded boxes represent actions, green hexagons the input of a phase, and blue trapezoids the output of a phase. This phase gets as input the \textit{Time Series}, which is prepared for the following phases. As forecasting methods, especially machine learning methods, struggle with changing variance and multiplicative effects within a time series~\cite{SugiyamaNonStationary}, the time series is transformed. More precisely, Telescope applies the \textit{Box-Cox Transformation} (see Section~\ref{foundations:sec:tstransformation}) to the \textit{Time Series}. We integrated this transformation step as it reduces both variance and multiplicative effects of the time series, leading to an improved forecast model~\cite{makridakis2018statistical,HA14}. For estimating the \textit{Transformation Parameter} of the Box-Cox transformation, we apply the method proposed by Guerrero~\cite{guerrero1993time} and restrict the parameter to values greater than or equal to zero. As the Box-Cox transformation can result in a logarithmic transformation, the time series has to be ``real'' positive ( $\forall t: y_t >0$). Consequently, the time series is shifted along the ordinate before the transformation if there is at least one value less than or equal to zero. 

In parallel to the transformation, Telescope performs the \textit{Frequencies Estimation}. In short, the main idea of this step is to retrieve the most dominant\footnote{By dominant, we mean the most common period such as days in a year.} frequencies from the input time series by applying a periodogram (see Section~\ref{foundations:sec:spectralanalysis}).

\subsection{Feature Extraction}
\label{approach:sec:featureextraction}

\begin{figure}[htb!]
	\centering
	\includegraphics[width=\linewidth]{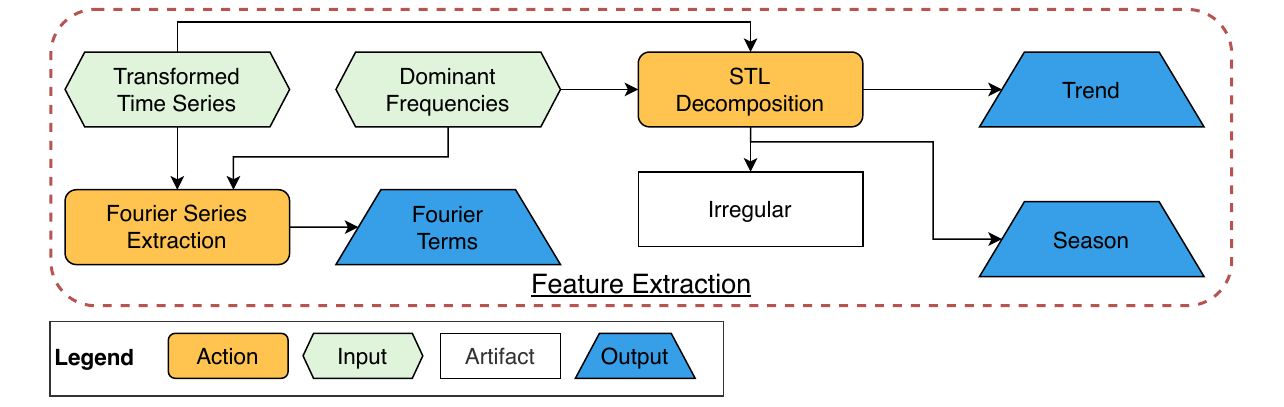}
	\caption{Feature extraction phase of Telescope.}
	\label{fig:tel:featureextraction}
\end{figure}

The second phase, \textit{Feature Extraction}, is depicted in Figure~\ref{fig:tel:featureextraction}. As input, this phase gets the \textit{Transformed Time Series} and the \textit{Dominant Frequencies} from the \textit{Preprocessing} phase. Based on the input, Telescope retrieves intrinsic time series features for tackling typical problems or difficulties that may occur during the modeling of a time series. The first difficulty is that time series may have multiple underlying seasonal patterns~\cite{HA14}. To this end, Telescope determines for each dominant frequency the associated \textit{Fourier Terms} (see Section~\ref{foundations:sec:fourierterms}) of the \textit{Transformed Time Series} for modelling the different patterns. More precisely, for each dominant frequency, a sine as well as a cosine with the period length of the corresponding frequency are retrieved from the time series.

Another problem are time series violating the stationary property (see Section~\ref{foundations:sec:stationarity}). Although most forecasting methods assume stationary time series~\cite{brockwell2016introduction}, many time series exhibit trend and/or seasonal and thus are non-stationary~\cite{adhikari2013introductory}. To handle the non-stationarity, the core idea of Telescope is to decompose the time series and then deal with each part separately. To this end, the \textit{Transformed Time Series} is split into its components \textit{Trend}, \textit{Season}, and \textit{Irregular} with STL (see Section~\ref{foundations:sec:tsdecomposition}). For the decomposition task, the most dominant frequency is used to specify the length of the seasonal pattern and the extraction of the pattern is set to periodic, that is, we assume that the seasonal pattern does not evolve over time. Although STL can only deal with an additive relationship between the components of a time series, it is not examined whether the time series follows an additive or multiplicative decomposition. More precisely, we assume that the Box-Cox transformation in the \textit{Preprocessing} phase has minimized or removed the multiplicative effects.

\subsection{Model Building}
\label{approach:sec:modelbuilding}

In the third phase called \textit{Model Building}, the model that reflects the time series is build on the inputs \textit{Transformed Time Series}, \textit{Trend}, \textit{Fourier Terms}, and \textit{Season}. To build a suitable model describing the time series, we apply machine learning for finding the relationship between the time series and the intrinsic features. In a time-critical scenario, that is, the forecast is required with a reliable time-to-result, Telescope implements XGBoost~\cite{chen2016xgboost} as regression-based machine learning method. We choose XGBoost since boosting tree algorithms are time-efficient, accurate, and easy to interpret~\cite{ke2017lightgbm}. In a scenario where the time-to-result is negligible, Telescope uses its recommendation system for selecting the most appropriate regression-based machine learning method for the given time series.

\subsubsection{Time-Critical Scenario}

\begin{figure}[htb!]
	\centering
	\includegraphics[width=\linewidth]{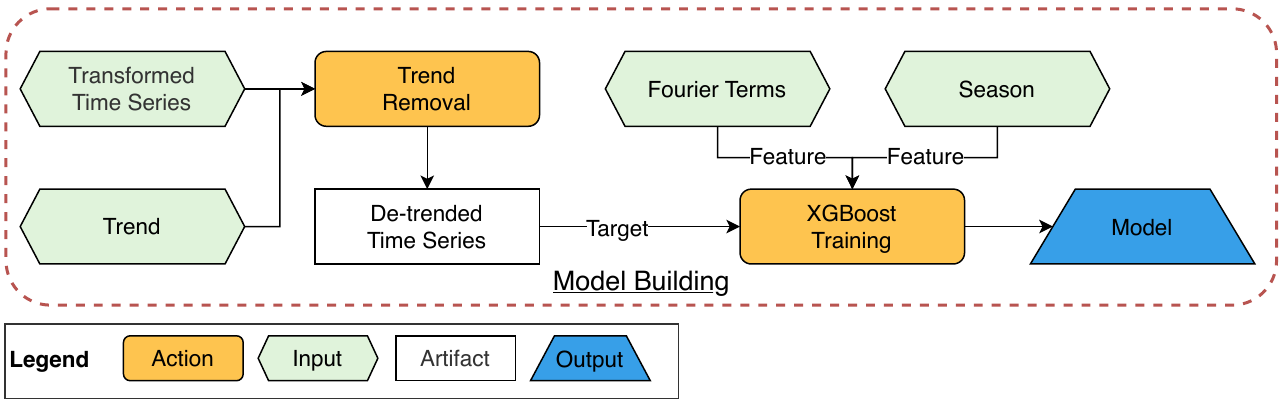}
	\caption{Model building phase of Telescope in a time-critical scenario.}
	\label{fig:tel:modelbuilding}
\end{figure}

Figure~\ref{fig:tel:modelbuilding} shows the \textit{Model Building} phase in a time-critical scenario. As a strong trend both increases the variance and violates stationarity, the trend introduces challenges for the model building. To this end, the first step is the removal of the trend from the time series. The resulting \textit{De-trended Time Series} is now trend-stationary. Although seasonality can also violate stationarity, machine learning methods are suitable for pattern recognition~\cite{dietterich2002machine}. Consequently, XGBoost learns during its training procedure how the \textit{De-trended Time Series} can be described by the intrinsic features \textit{Fourier Terms} and \textit{Season}. Note that the irregular part of the time series is not explicitly considered a feature to reduce the model error and later the forecast error. That is, the machine learning method learns the irregular part as the difference that is missing to fully recreate the de-trended time series. 

\subsubsection{Non-Time-Critical Scenario}

\begin{figure}[htb!]
	\centering
	\includegraphics[width=\linewidth]{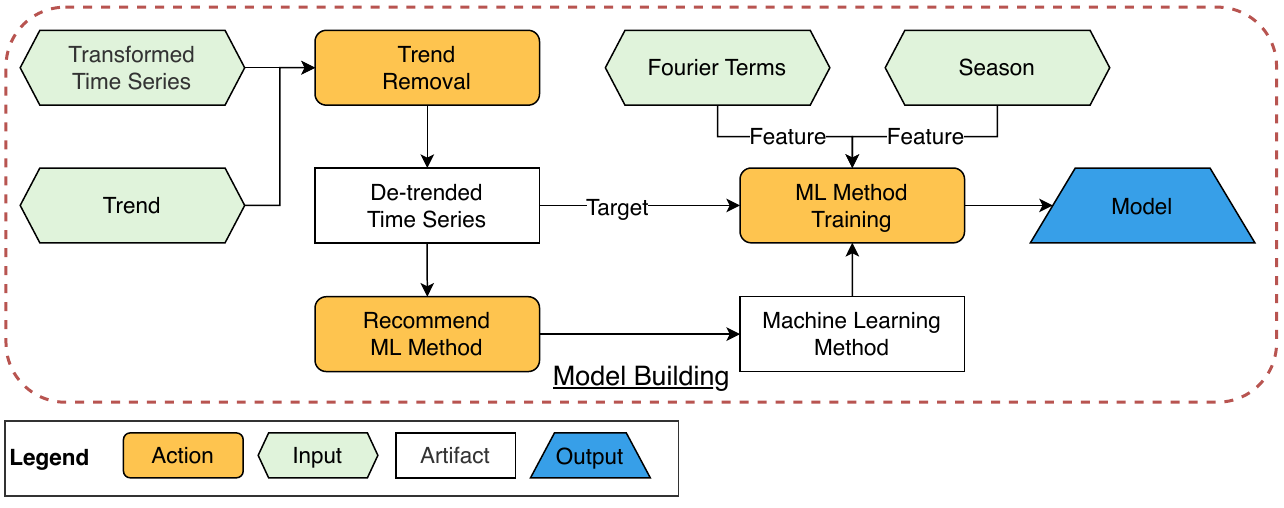}
	\caption{Model building phase of Telescope in a non-time-critical scenario.}
	\label{fig:tel:modelbuilding_offline}
\end{figure}

The \textit{Model Building} phase in a non-time-critical scenario, which is depicted in Figure~\ref{fig:tel:modelbuilding_offline}, is identical to the phase in a time-critical scenario, with the exception that no fixed machine learning method is used. In other words, the machine learning method is selected based on the time series. To this end, the \textit{De-trended Time Series} is passed to the recommendation system that extracts characteristics of the time series. Based on these characteristics, the most suitable machine learning method is selected and used to learn how the intrinsic features can describe the time series. The details of the recommendation system are described in Section~\ref{approach:sec:recommendation}.

\subsection{Forecasting}
\label{approach:sec:forecasting}

\begin{figure}[htb!]
	\centering
	\includegraphics[width=\linewidth]{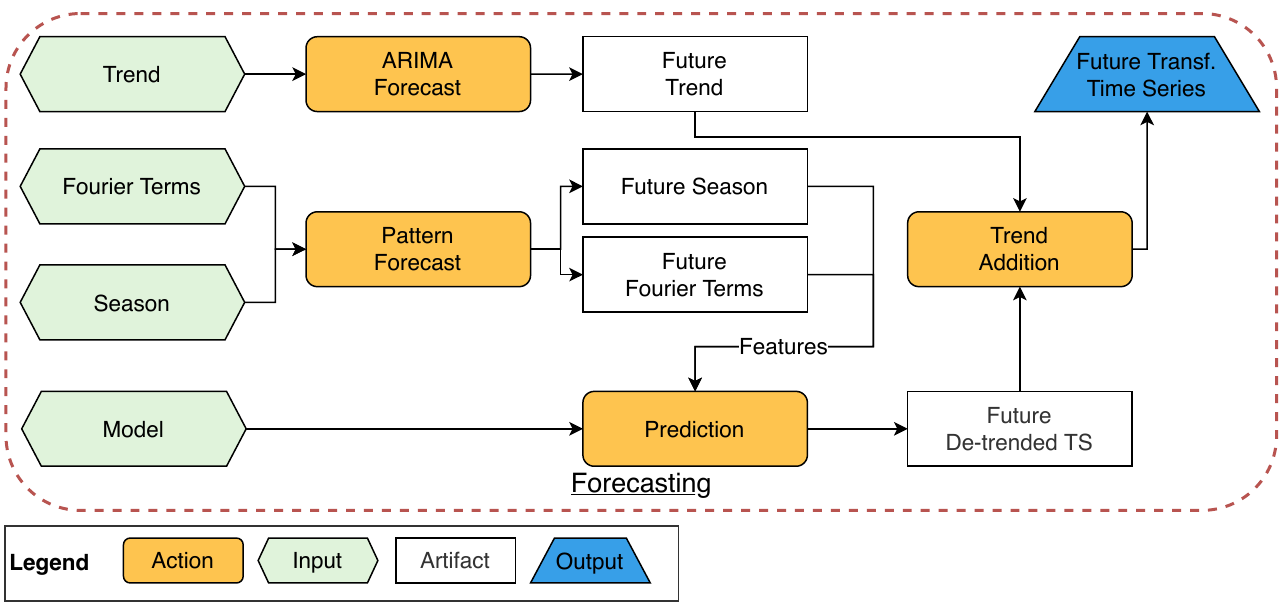}
	\caption{Forecasting phase of Telescope.}
	\label{fig:tel:forecasting}
\end{figure}

In the \textit{Forecasting} phase, which is illustrated in Figure~\ref{fig:tel:forecasting}, the different components are forecast separately and then, the future time series is assembled. To this end, this phase gets as input the \textit{Trend}, \textit{Fourier Terms}, \textit{Season} as well as the \textit{Model}. As the \textit{Season} and the \textit{Fourier Terms} are recurring patterns per definition, they can be merely continued. More precisely, the seasonal pattern and each Fourier terms is forecast for the time $n+k$ as follows
\begin{align}
\hat{y}_{n+k|n} := y_{n + k - m \cdot ( \lfloor\frac{k - 1}{m}\rfloor + 1)},
\end{align}
where $y_t$ is the observations at time $t$, $n$ the number of historical observations, $m$ the length of the associated period, and the forecast horizon $k$ being a positive integer. The resulting \textit{Future Season} and \textit{Future Fourier Terms} are used as features in conjunction with the \textit{Model} for predicting the future de-detrended time series. More precisely, the machine learning method regresses a new value of the \textit{Future De-trended TS} for each point in time of the forecast based on the corresponding values of the features.

In parallel to the forecast of the recurring patterns, the trend is also forecast. Since the \textit{Trend} contains no recurring patterns, an advanced forecasting method is required to forecast the \textit{Future Trend}. To this end, we apply ARIMA as it is able to estimate the trend even from a few points. More precisely, we apply auto.arima\footnote{Auto.arima conducts a search over possible sARIMA and ARIMA models.}~\cite{forecastPackageRpaper} that automatically finds the most suitable model parameters for a time series. After the trend is forecast, the last step of this phase is assembling the forecast of the time series. To this end, the \textit{Future De-trended TS} and the \textit{Future Trend} are summed up. 

\subsection{Postprocessing}
\label{approach:sec:postprocessing}

\begin{figure}[htb!]
	\centering
	\includegraphics[width=\linewidth]{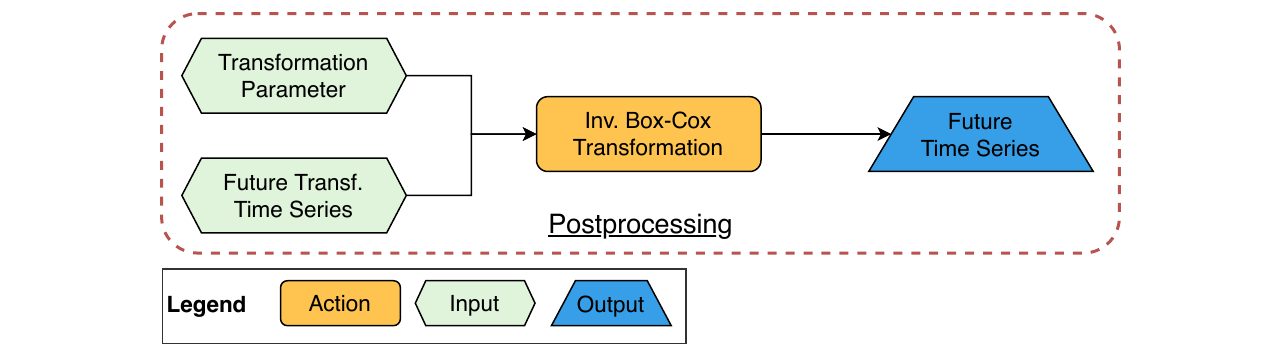}
	\caption{Postprocessing phase of Telescope.}
	\label{fig:tel:postprocessing}
\end{figure}

The last phase of Telescope is called \textit{Postprocessing} and its workflow is depicted in Figure~\ref{fig:tel:postprocessing}. As the name suggests, this phase is the counterpart of the \textit{Preprocessing} step. More precisely, it gets as input the \textit{Future Transf. Time Series} from the \textit{Forecasting} phase and the \textit{Transformation Parameter} from the \textit{Preprocessing} phase. As the time series was adjusted with the Box-Cox transformation, the \textit{Future Transformed Time Series} has to be re-transformed with the identical transformation parameter from the \textit{Preprocessing} phase. To this end, the inverse Box-Cox transformation with the \textit{Transformation Parameter} is applied to the \emph{Future Transformed Time Series}. If the time series was shifted along the vertical axis in the first phase, the time series also has to be moved back. Finally, the forecast of the original time series is returned. 

\subsection{Fallback for Non-Seasonal Time Series}
\label{approach:sec:fallback}

Telescope's core idea is to detect recurring patterns within a time series and use this information to retrieve intrinsic features. However, if Telescope has to forecast a non-seasonal time series, the normal workflow cannot be used as the time series lacks recurring patterns. Moreover, the integrated STL also requires a seasonal pattern to decompose the time series. Consequently, Telescope requires a fallback for non-seasonal time series. That is, if the \textit{Preprocessing} phase returns only the frequency of 1 as dominant frequency, the \textit{Feature Extraction}, \textit{Model Building}, and \textit{Forecasting} phase are omitted and the fallback is executed instead. The fallback receives as input the transformed time series from the \textit{Preprocessing} phase on which a non-seasonal ARIMA model is trained.  Then, the forecast is performed and the future transformed time series is forwarded to the \textit{Postprocessing} phase, in which the time series is treated as described in Section~\ref{approach:sec:postprocessing}.

\subsection{Recommendation System for Machine Learning Method}
\label{approach:sec:recommendation}

According to the ``No-Free-Lunch Theorem'', which -- simply spoken -- states that each method has its advantages and weaknesses depending on the specific use case, it is inadvisable to rely only on one particular method. Typically, the choice of the optimal forecasting method is based on expert knowledge, which can be expensive, may have a subjective bias, and may take a long time to deliver results. Consequently, we automate the selection of the best regression-based machine learning method for a given time series. More specifically, we employ a recommendation system based on \emph{meta-learning} to select the most appropriate method based on time series characteristics. In the meta-learning context, the methods, which are available for selection, are referred to as \emph{base-level methods} and characteristics on which the selection is based are called \emph{meta-level attributes}.

\subsubsection{Problem Formalization}
\label{approach:sec:metalearning}

Following the idea of the \emph{algorithm selection problem}~\cite{rice1976algorithm} and its formal description~\cite{smith2009cross}, the \emph{regression-based machine learning method selection problem} that arises for Telescope can be formally defined as: For a given time series $y \in Y$, with characteristics $f(y) \in F$, find the selection mapping $S(f(y))$ into the algorithm space $A$, such that the selected algorithm $a \in A$ minimizes the forecast error measure $m(a(y)) \in M$. In this formulation, the problem space $Y$ represents a set of time series; the feature space $F$ contains measurable time series characteristics of each instance of $Y$, calculated by a deterministic extraction procedure; the algorithm space $A$ is the set of all considered regression-based machine learning methods; the performance space $M$ represents the mapping of each algorithm to the forecast error measure.

\subsubsection{Meta-Level Attributes}
\label{approach:sec:metalevelattributes}

To have an accurate recommendation system for choosing the most appropriate regression-based machine learning method for a given time series, a sound set of characteristics, which describe the time series, is required. To this end, the considered time series characteristics originate from different sources: statistical measures of a time series (S1--S6) as well as characteristics proposed in a former paper~\cite{BaZuGrScHeKo-ICPE20-Seasonal-Forecast} (B1--B8), proposed by Lemke and Gabrys~\cite{lemke2010meta} (L1--L4), and proposed by Wang et al. \cite{Wang20092581} (W1--W6).  In contrast to the work of Wang et~al.\cite{Wang20092581}, we use the raw values of the characteristics to avoid arbitrary normalization factors. The time series characteristics applied to the meta-learning approach are listed in Table~\ref{tab:metalevel:attributes}. Note that these characteristics are only a subset of the original set, selected by, for example, correlation analysis.  

\begin{table}[htb!]
	\centering
	\caption{Meta-Level Attributes}
	\resizebox{\columnwidth}{!}{
	\begin{tabular}{lp{8cm}}
		\toprule
		\textbf{ID}              & \textbf{Description} \\ \midrule
		S1 & The \emph{frequency} specifies the length of the most dominant recurring pattern (e.g., 365 daily observations in a yearly pattern) within the de-trended time series.              \\
	    S2 & The \emph{length} counts the total number of observations included in the de-trended time series. \\
        S3 & The \emph{standard deviation} measures the amount of variations within the de-trended time series. \\
        S4 & The \emph{skewness} reflects the symmetry of the value distribution of the de-trended time series. \\
        S5 & The \emph{irregular skewness} quantifies the skewness of the irregular part of the time series. \\
        S6 & The \emph{irregular kurtosis} reflects the tailedness of the value distribution of the irregular part of the de-trended time series.  \\ 
        B1 & The \emph{mean period entropy} quantifies the regularity and unpredictability of fluctuations of the de-trended time series. For this purpose, the approximate entropy of each full period is calculated and then averaged. \\
    	B2 & The \emph{coefficient of entropy variation} describes the standardized entropy distribution over all periods. \\
    	B3 & The \emph{mean cosine similarity} states how similar all periods are to each other. The similarity is expressed with the average cosine similarity of each full pair of periods. \\
    	B4 & The \emph{sinus approximation} quantifies how well the seasonal pattern of the time series can be approximated by a sinus wave. To this end, the Durbin-Watson statistic~\cite{durbin1950testing} is used to measures the autocorrelation of the resulting fitted errors. \\ 
    	L1 & The \emph{2$^{nd}$ frequency} states the second most dominant frequency of the de-trended time series. \\
    	L2 & The \emph{3$^{rd}$ frequency} refers to the third most dominant frequency of the de-trended time series. \\
    	L3 & The \emph{maximum spectral value} specifies the maximum spectral value of the periodogram applied to the de-trended time series. \\
    	L4 & The \emph{number of peaks} reflects how many strong recurring patterns the de-trended time series has. More precisely, the number of peaks in the periodogram that are at least 60\% of the maximum value is counted.  \\
    	W1 & The \emph{strength of seasonal component} measures the degree of the seasonality within the de-trended time series. \\ 
    	W2 & The \emph{serial correlation} describes the correlation of the de-trended time series with itself to an earlier time. \\
    	W3 & The \emph{irregular serial correlation} states the serial correlation of the irregular part of the time series. \\
    	W4 & The \emph{non-linearity} quantifies the degree of the non-linearity of the de-trended time series.  \\
    	W5 & The \emph{irregular non-linearity} reflects the degree of the non-linearity of the irregular part of the time series. \\
    	W6 & The \emph{self-similarity} measures how similar the de-trended time series is to a part of itself. \\ \bottomrule
	\end{tabular}
	}
	\label{tab:metalevel:attributes}
\end{table}

\subsubsection{Base-Level Methods}
\label{approach:sec:baselevelmethods}

We only consider machine learning methods to learn how the de-trended time series can be described with intrinsic time series features. 
Telescope implements seven regression-based machine learning methods\footnote{The selection contains the seven best-performing methods during preliminary experiments.} as the base-level methods: (i) \textit{CART}~\cite{breiman1984classification} is a regression tree that recursively partitions the data set. To prevent the tree from becoming too large, the tree is automatically pruned. 
(ii) \textit{Cubist} is a rule-based regression method that builds upon the \emph{M5} model tree \cite{quinlan1993combining}. The rules are arranged hierarchically, resulting in a tree where each leaf node represents a multivariate linear regression model. 
(iii) \textit{Evtree}~\cite{grubinger2011evtree} implements an evolutionary algorithm for constructing a regression tree that splits the data so that each partition decision is globally optimal. 
(iv) \textit{NNetar}~\cite{forecastpackage} is a feed-forward neural network consisting of one hidden layer and is trained with lagged values of the time series. The number of nodes in the hidden layer and the lags are automatically determined. 
(v) \textit{Random forest}~\cite{breiman2001random} is an ensemble method comprising multiple regression trees. Each tree is built independently of the others, and for each split, only a random subset of the features is considered.
(vi) \textit{SVR}~\cite{drucker1997support} uses the same principles as SVM. More precisely, a set of hyperplanes is constructed for separating the data. If the data is not linear in its input space, the values are mapped into a higher dimensional feature space.
(vii) \textit{XGBoost} is an ensemble method consisting of multiple regression trees and uses gradient tree boosting, i.e., each tree grows with knowledge from the last trained tree.

\subsubsection{Rule Generation Approach}
\label{approach:sec:selectionapproaches}

\begin{figure}[htb!]
	\centering
	\includegraphics[width=\linewidth]{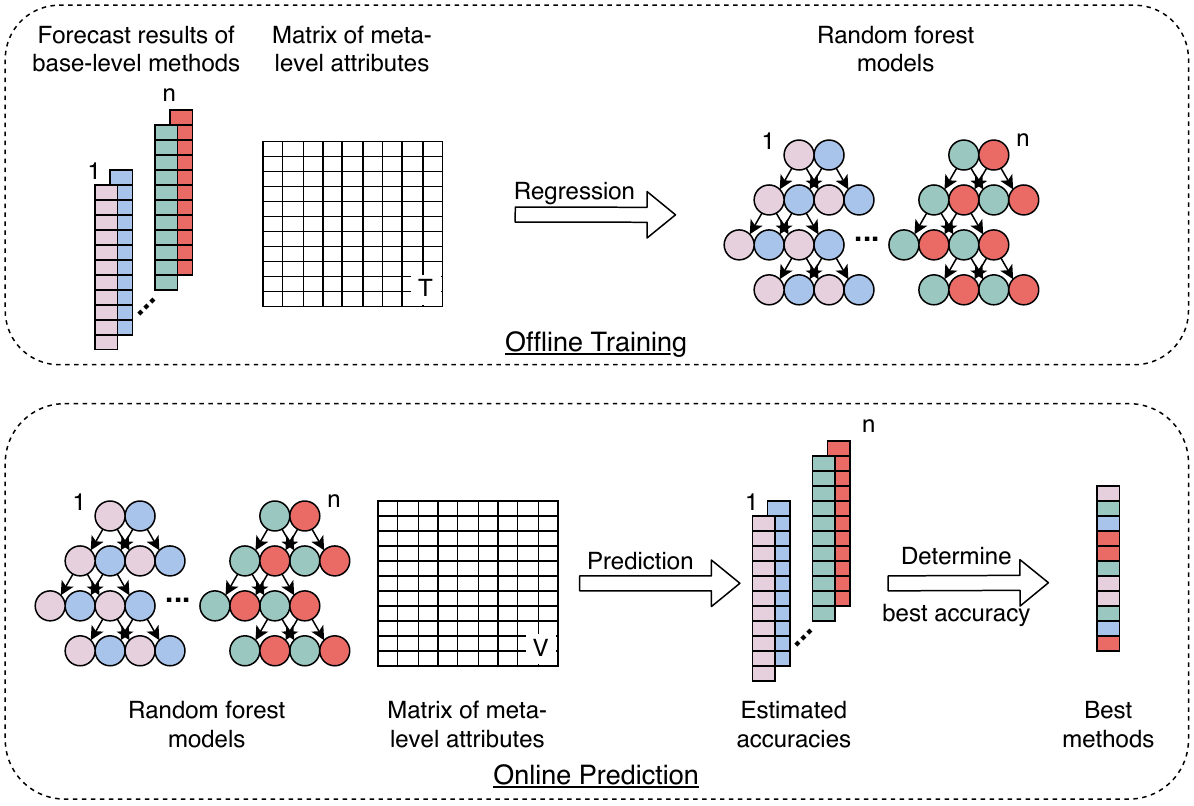}
	\caption{Schematic process of the rule generation.}
	\label{fig:rec:reg}
\end{figure}

To recommend the most appropriate regression-based machine learning method for a given time series, we propose a regression-based rule generation approach, as illustrated in Figure~\ref{fig:rec:reg}. The idea is to apply a random forest\footnote{The choice of random forest for the recommendation is based on the extensive experiments in which these combinations yielded the best results.} for learning how meta-level attributes (i.e., time series characteristics) can be mapped to the performance of the base-level methods (i.e., regression-based machine learning methods). 

In detail, for each base-level method, a random forest is trained. Each model estimates the forecast accuracy for a given time series. Then, the method that has the best accuracy based on the estimations is selected. More precisely, the approach calculates for each base-level method and each time series how worse it is compared to the method with the best forecast accuracy. Then, the method that has the best accuracy based on the estimations is selected. More precisely, the approach calculates for each base-level method and each time series how worse it is compared to the method with the best forecast accuracy. We reflect this deterioration with the \emph{forecast accuracy degradation} that can be calculated as 
\begin{align}
\vartheta_i := \frac{\epsilon_i}{min(\epsilon_1,\ldots,\epsilon_n)},
\end{align}
where $\epsilon_i$ is the forecast accuracy of the $i$-th method and $n$ is the number of considered methods. The values of $\vartheta_i$ lie in the interval $[1, \infty)$, where 1 indicates that this method has the best forecast accuracy. After the calculation of the degradation, a random forest is used as regressor for each base-level method, where the meta-level attributes are the features and the forecast accuracy degradation vector of the respective method is the target. In other words, the regression task leads to $n$ random forest models, each reflecting the estimate of the forecast accuracy degradation compared to the best method for a given time series. After the training, the meta-level attributes of new time series can be fed to each of the random forest models. Based on these attributes, each model estimates the forecast accuracy degradation vector. Then, the base-level methods with the lowest estimated forecast accuracy degradation are returned for each time series.

\subsubsection{Offline Training}
\label{approach:sec:offlinetraining}

\begin{figure}[htb!]
	\centering
	\includegraphics[width=0.75\linewidth]{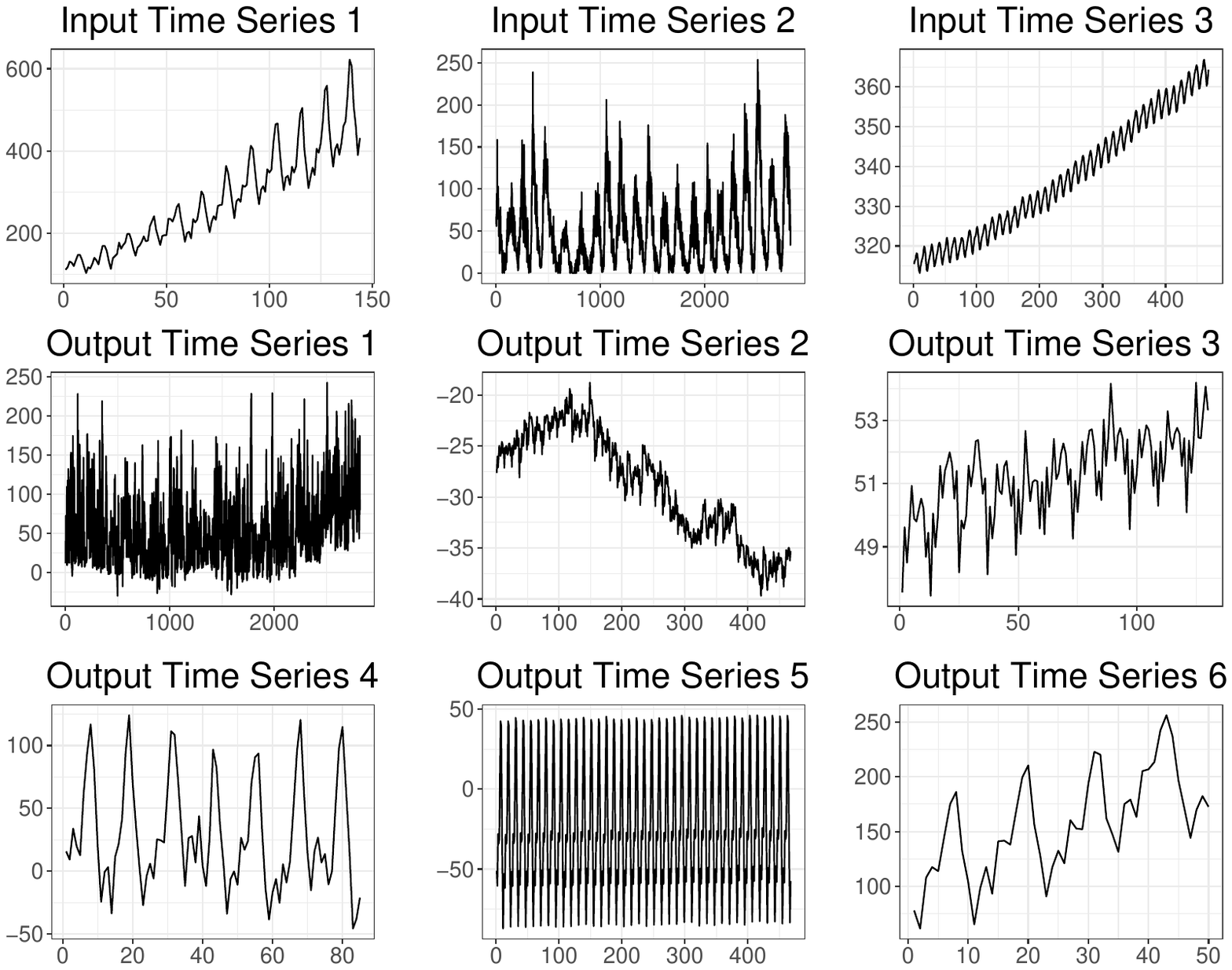}
	\caption{Example of six generated time series.}
	\label{fig:newtimeseries}
\end{figure}

\begin{figure}[htb!]
	\centering
	\includegraphics[width=\linewidth]{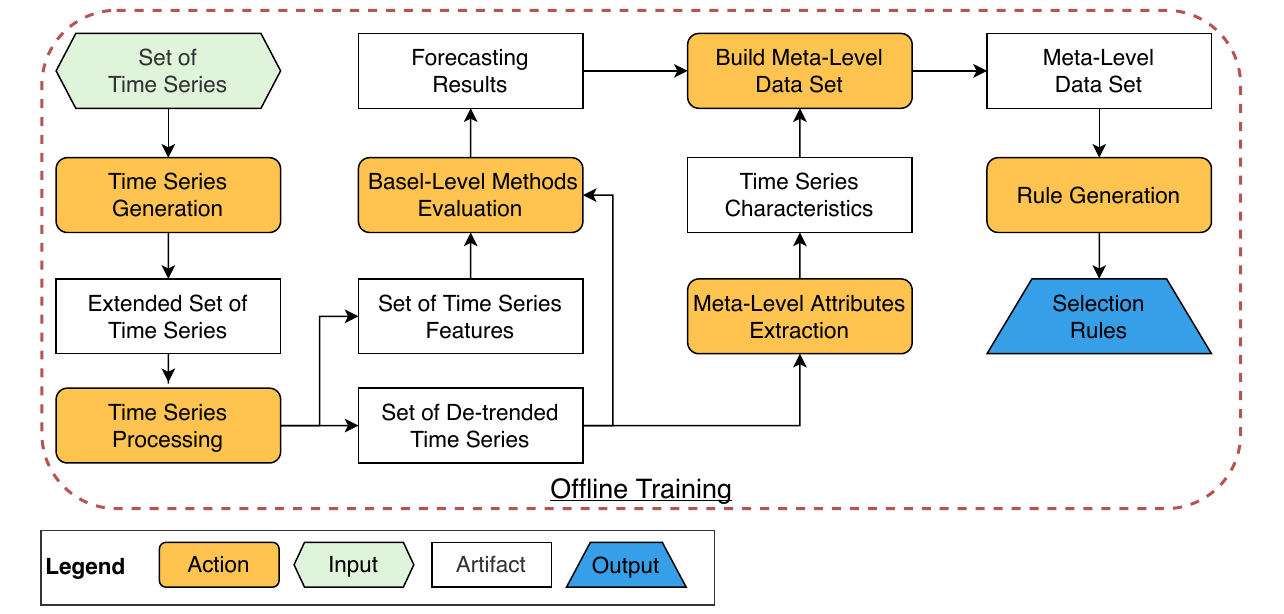}
	\caption{Offline training phase of Telescope.}
	\label{fig:rec:offlinetraining}
\end{figure}

In the \emph{Offline Training} phase, which is depicted in Figure~\ref{fig:rec:offlinetraining}, the rules for recommending a specific method based on time series characteristics are learned or updated either when Telescope is started or when no forecast is currently being conducted. To this end, this phase gets a \textit{Set of Time Series} as input. To retrieve precise rules for the recommendation of the most appropriate regression-based machine learning method for a given time series, a set of time series which may be similar to this time series is required. Therefore, Telescope generates $n$ new time series based on the initial set of time series in this phase. The main idea is to decompose all time series in the set into components (trend, season, and irregular part) and then, build randomly new time series based on these components. An example output of this time series generation is illustrated in Figure~\ref{fig:newtimeseries}.

As the machine learning method has to learn how the de-trended time series can be described by the intrinsic features, each time series in the \textit{Extended Set of Time Series}, which comprises the initial time series and the newly generate time series, has to be de-trended. For this purpose, each time series is transformed with the Box-Cox transformation, de-trended, and the intrinsic features are extracted.  

In the \textit{Base-Level Methods Evaluation} step, each base-level method is trained and evaluated on every de-trended time series and its associated intrinsic features (Fourier terms and seasonal pattern). To this end, the time series is split into history (the first 80\% of the time series) and in future/test (the remaining 20\%). In parallel, the time series characteristics (i.e., meta-level attributes) of each de-trended time series are extracted. The \textit{Forecasting results} from the \textit{Base-Level Methods Evaluation}, in this case the forecast error based on sMAPE (see Section~\ref{results:sec:setup:measures}), and the \textit{Time Series Characteristics} are used to form the \textit{Meta-Level Data Set}. This data set is used in the \textit{Rule Generation} step to retrieve the rules for the recommendation of the best suited method.

\subsection{Assumptions and Limitations}
\label{approach:sec:limitations}

In the development of Telescope, we limit ourselves to univariate time series. In fact, correlated/external data can be used for each time series to improve forecast accuracy. However, the selection and preprocessing of such additional information require domain knowledge. In other words, this knowledge about domain-specific feature engineering cannot yet be fully automated. Consequently, our method would have to be tailored to a specific domain and, therefore, contradict the goal of a generic forecasting approach. Besides this limitation, Telescope has the following assumptions that are either based on the integrated tools or design decisions: 
(i)~The time series must not contain missing values, and each value must be numeric. Indeed, missing values can be interpolated. However, if the gaps are large or frequent, the values must be treated with caution, as they can affect the model's accuracy. 
(ii)~For the recommendation of the most appropriate regression-based machine learning method, Telescope assumes that the time series to be forecast and the time series that are used for the training (i.e., the initial time series set plus the generated time series) originate from the same population and therefore, have a similar distribution of time series characteristics. In other words, Telescope assumes the rules that are retrieved based on the training set can also be applied to new time series.
(iii)~Usually, many systems are driven by human interactions. In other words, the time series produced or observed by these systems are subject to human habits (e.g., day/night phases) and, therefore, seasonal. Consequently, Telescope assumes that the found frequencies within time series are multiples of natural frequencies.
(iv)~We expect that the seasonal pattern does not evolve over time.

\section{Forecasting Method Competition}
\label{sec:results}

To compare Telescope with recent hybrid forecasting methods, we compare Telescope to seven competing methods using a forecasting benchmark. We start with the description of the experiments. Then, we asses the performance of Telescope and the competing methods in Section~\ref{results:sec:telescope}. A more detailed investigation of the forecasting methods takes place
in Section~\ref{results:sec:telescope:sec:detailed}. Lastly, we summarize the results and discuss threats to validity in Section~\ref{results:sec:telescope:sec:threats}.

\subsection{Experimental Description}

In this evaluation, we investigate both the pure Telescope approach and Telescope using the recommendation system. For the training of the recommendation system, we assembled a broad data set of  150  real-world  time  series  with  varying  length  and frequencies. Note that the applied benchmark and this training data set consist of different time series. Moreover, Telescope augment the training data set consisting of 150 time series to 10,000. In the following, we refer to Telescope using the recommendation system as \emph{Telescope$^*$}.

\subsubsection{Applied Forecasting Benchmark}
\label{results:sec:benchmark}
For benchmarking the forecasting methods, we apply \textit{Libra}~\cite{bauer2021benchmark}. Libra is a benchmark for time series forecasting methods comprising four different use cases, each covering 100 heterogeneous time series taken from different domains. In the following evaluation, we benchmark all methods with Libra across all use cases (i.e., 400 time series). Moreover, the evaluation type of Libra was set to multi-step-ahead forecasts. That is, each time series was divided into history (first 80\% of the time series) and test (remaining 20\% of the time series). Based on the history, each method learned a model that was used for forecasting future values of the time series (i.e., the test part) at once with a single execution. This forecasting procedure (i.e., receiving the time series, estimating the parameters, building the model, and forecasting the time series) was repeated ten times for each time series. Consequently, the reported measures were determined on the average values of each time series.

\subsubsection{Competing Methods}
For having a broad and representative forecasting method competition, we compare different methods from different fields. The competitors can be grouped into three categories: (i) ``classical'' time series forecasting methods, (ii) regression-based machine learning methods, and (iii) hybrid forecasting methods (i.e., taking advantage of at least two methods). For the first and second groups, we consider their best performing representative on the benchmark (i.e., sARIMA and XGBoost)~\cite{bauer2021benchmark}. As representatives for the hybrid methods, we investigate recent existing approaches. The considered seven forecasting methods are listed and briefly described below:
(i)~\textit{BETS} decomposes the time series into the components trend, season, and irregular. Different versions of the irregular part are then simulated, resulting in different versions of the original time series. Then, these versions are forecast separately by ETS~\cite{HyndmanETS}, and the final forecast is the median of all forecasts.
(ii)~\textit{ES-RNN} is a hybrid forecasting method based on time series decomposition developed by Uber. The basic idea is to de-seasonalize the time series using exponential smoothing and to use a neural network for extrapolation of the time series. 
(iii)~\textit{FFORMS} is based on forecasting method recommendation. More precisely, a random forest is applied as a classifier to map the most appropriate forecasting method (ETS, NNetar, sARIMA, sNa{\"i}ve~\cite{forecastpackage}, TBATS~\cite{de2011forecasting}, and Theta~\cite{assimakopoulos2000theta}) to a specific time series described by a set of time series characteristics. 
(iv)~\emph{Hybrid}\footnote{Hybrid forecasting method: \url{https://cran.r-project.org/web/packages/forecastHybrid/vignettes/forecastHybrid.html}} performs an ensemble forecast. The considered methods comprise ETS, NNetar, Theta, sARIMA, and TBATS. For the ensemble forecast, each method performs a forecast, and the final forecast is the average of these forecasts.
(v)~\textit{Prophet} is a hybrid forecasting method based on decomposition developed by Facebook. The time series is decomposed into the components trend, season, holiday, and error. Each component is forecast by a different approach. Then, the forecast parts are assembled to form the final forecast. 
(iv)~\textit{sARIMA}\footnote{In the experiments, we use auto.arima~\cite{HA14} to find the most suitable model for the time series automatically.} extends the ARIMA model~\cite{box1970time} by adding a seasonal counterpart to each component (autoregressive model for the past values, moving average for the past forecast errors, and time series differencing for stationarity).
(vii)~\textit{XGBoost} is an ensemble of decision trees based on gradient tree boosting, that is, the trees are growing sequentially with knowledge from their preceding tree. To reduce overfitting, XGBoost applies regularization objects, shrinkage, and feature subsampling.

As input for the forecasting task, sARIMA, BETS, NNetar, and Hybrid received the time series and the respective frequency. Prophet also needs the timestamps of the time series. XGBoost received the time series and a synthetic seasonal pattern (a vector with the indices modulus the frequency) as input. FFORMS was trained on the M4- and M3-Competition~\cite{makridakis2000m3} and received as input also the time series and the respective frequency. In contrast to the other methods, ES-RNN requires, besides the time series and frequency, a set of time series. According to the workflow of Libra, ES-RNN got the same time series several times. Note that we used all methods "out-of-the-box" since the results of the M3-Competition have shown that the methods were kept simple and that complex models do not necessarily perform better~\cite{makridakis2000m3}. That is, there was no parameter tuning, and the methods were used with their default settings. Recall the ``No-Free-Lunch Theorem'', stating that improving a method for one aspect leads to deterioration in performance for another aspect. Moreover, also the techniques deployed in Telescope were used with their default settings.

\subsubsection{Considered Measures}
\label{results:sec:setup:measures}

To compare and quantify the performance of the different forecasting methods, we report the symmetrical mean absolute percentage error (sMAPE) and a normalized time-to-result\footnote{The time-to-result for a time series reflects the duration in which the forecasting method receives the time series, estimates the parameters, creates the model, and performs the forecast.}. The sMAPE is defined as following where $n$ is the forecast length, $y_t$ the actual value, and $f_t$ the forecast value: 
\begin{align}
    \text{sMAPE} &:= \frac{200\%}{n} \sum_{t=1}^{n} \lvert\frac{y_t - f_t}{y_t + f_t}\rvert.
\end{align}
In the following, the measures $\overline{e}$, $\tilde{e}$, and $\sigma_{e}$ reflect the average error, median error, and the standard deviation of the sMAPE, while the measures $\overline{t}_{N}$, $\tilde{t}_{N}$, and $\sigma_{t_{N}}$  reflect the average time, median time, and the standard deviation of the time-to-result normalized by the time required by a na{\"i}ve forecast\footnote{The na{\"i}ve forecast needed on average 0.01 seconds per forecast.}.

\subsection{Benchmarking the Telescope Approach}
\label{results:sec:telescope}

\begin{figure}[htb!]
	\centering
	\includegraphics[width=0.75\linewidth]{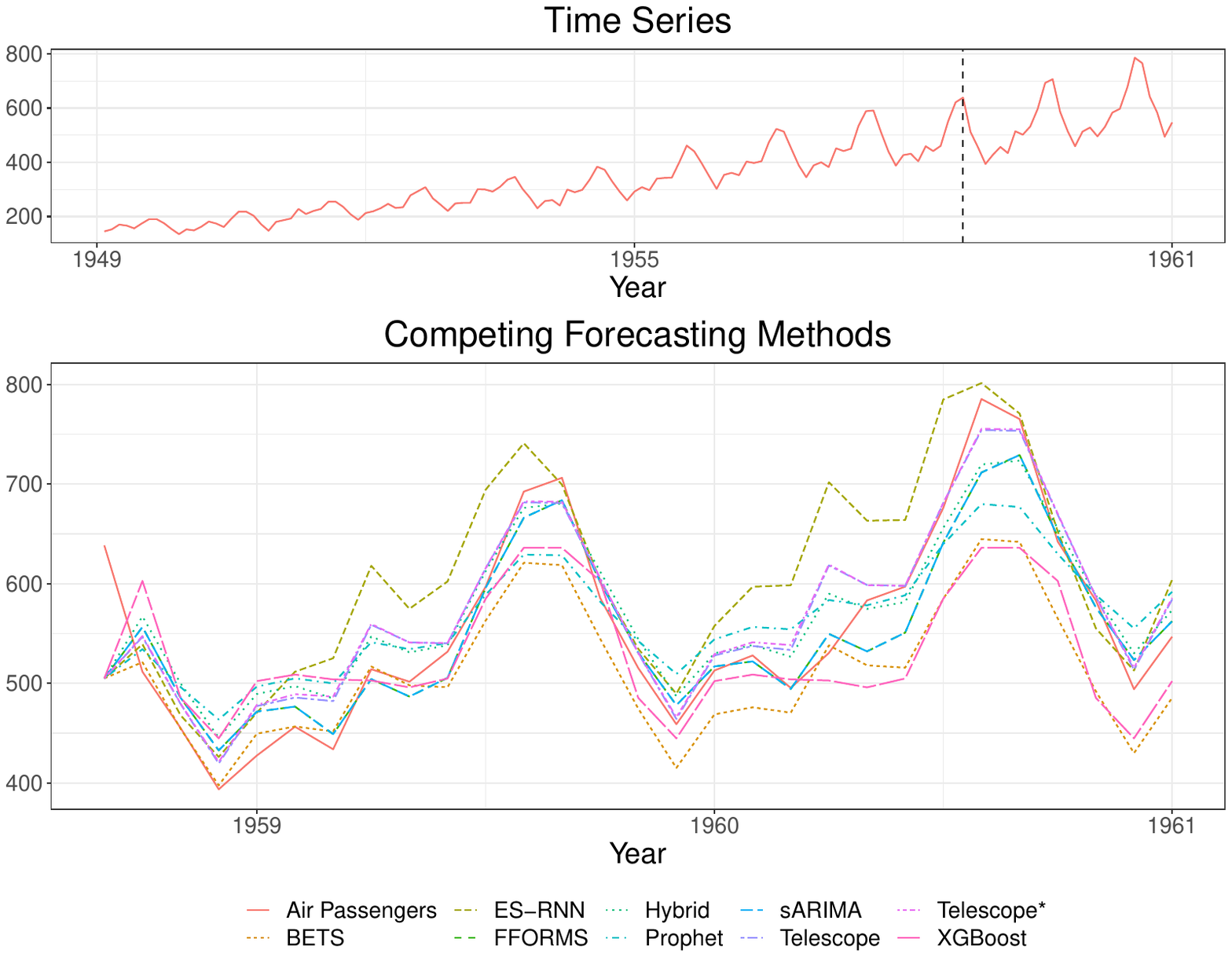}
	\caption{Forecasts for all methods in competition on the airline passengers time series.}
	\label{fig:diffforecasts}
\end{figure}

In this section, we compare Telescope against the competing forecasting methods. As an example, Figure~\ref{fig:diffforecasts} shows the different forecasting methods on the AirPassengers~\cite{box2015time} time series. This time series, which is often used as a baseline, shows that all forecasting methods are correctly configured and perform reasonable forecasts. Therefore, we investigate the results (i.e., forecast error and time-to-result) of these methods over all time series in the following.

\begin{table*}[htb!]
	\centering
	\caption{Forecast error and time-to-result comparison on all time series.}
		\begin{tabular}{lrrrrrrrrr}
			\textbf{Measures}               & \textbf{BETS}    & \textbf{ES-RNN}  & \textbf{FFORMS}  & \textbf{Hybrid}  & \textbf{Prophet} & \textbf{sARIMA}  & \textbf{Telescope} & \textbf{Telescope$^*$}                        & \textbf{XGBoost} \\ \midrule
			$\overline{e}$ [\%] & 25.52            & 47.87            & 21.15            & 27.89            & 35.56            & 20.63            & 19.95              & \textbf{19.46}                                & 23.85            \\
			$\sigma_{e}$ [\%]   & 34.25            & 66.99            & 36.87            & 89.76            & 2.30$\cdot 10^2$ & 35.63            & 31.35              & \textbf{27.96}                                & 34.67            \\
			$\overline{t}_{N}$             & 6.14$\cdot 10^4$ & 1.61$\cdot 10^6$ & 5.23$\cdot 10^5$ & 1.09$\cdot 10^6$ & 2.04$\cdot 10^3$ & 8.48$\cdot 10^5$ & 1.43$\cdot 10^2$   & 3.21$\cdot 10^3$          & \textbf{6.73}    \\
			$\sigma_{t_{N}}$               & 1.07$\cdot 10^6$ & 4.92$\cdot 10^6$ & 7.93$\cdot 10^6$ & 8.08$\cdot 10^6$ & 6.15$\cdot 10^3$ & 9.08$\cdot 10^6$ & 98.67              & 1.33$\cdot 10^4$         & \textbf{15.59}   \\ \bottomrule
		\end{tabular}
	\label{tab:telescope:average}
\end{table*}

Table~\ref{tab:telescope:average} shows the performance for all competing methods in competition averaged over all time series of all use cases. Each row represents a measure, each column a method, and the best values (the lower, the better) are highlighted in bold. The most accurate forecasting method is Telescope$^*$ (19.46\%) followed by Telescope (19.95\%) and sARIMA (20.63\%). By far, the fastest method is XGBoost (6.73). Telescope (1.43$\cdot 10^2$) has the second-lowest time-to-result while Telescope$^*$ is, on average, ten times slower. Although sARIMA has the third-lowest forecast accuracy, it is, on average, almost 6000 times shower than Telescope and almost 300 times slower than Telescope$^*$. More precisely, the maximum actual forecast time of sARIMA for a time series was 465,574 seconds, which corresponds to almost 5.5 days.

We also investigate the variation of the forecast error and the time-to-result as a crucial property. The lowest standard deviation regarding $\overline{e}$ is shown by Telescope$^*$ (27.96\%) followed by Telescope (31.35\%) and BETS (34.25\%). In terms of the time-to-result, XGBoost has the lowest variation (15.59) followed by Telescope (98.67). 

\begin{figure}[htb!]
	\centering
	\includegraphics[width=0.75\columnwidth]{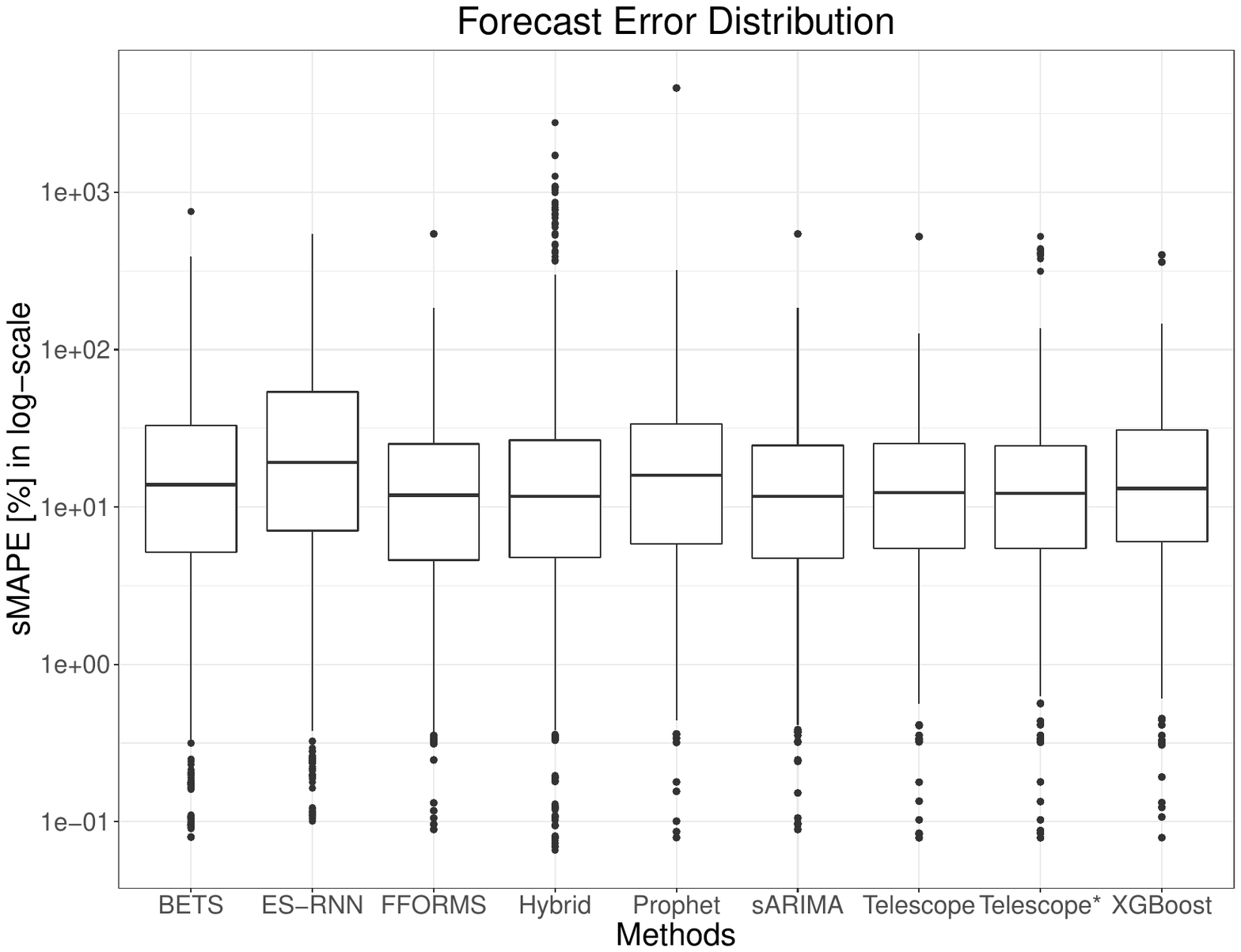}
	\caption{Forecast error distribution.}
	\label{fig:errordis}
\end{figure}

\begin{figure}[htb!]
	\centering
	\includegraphics[width=0.75\columnwidth]{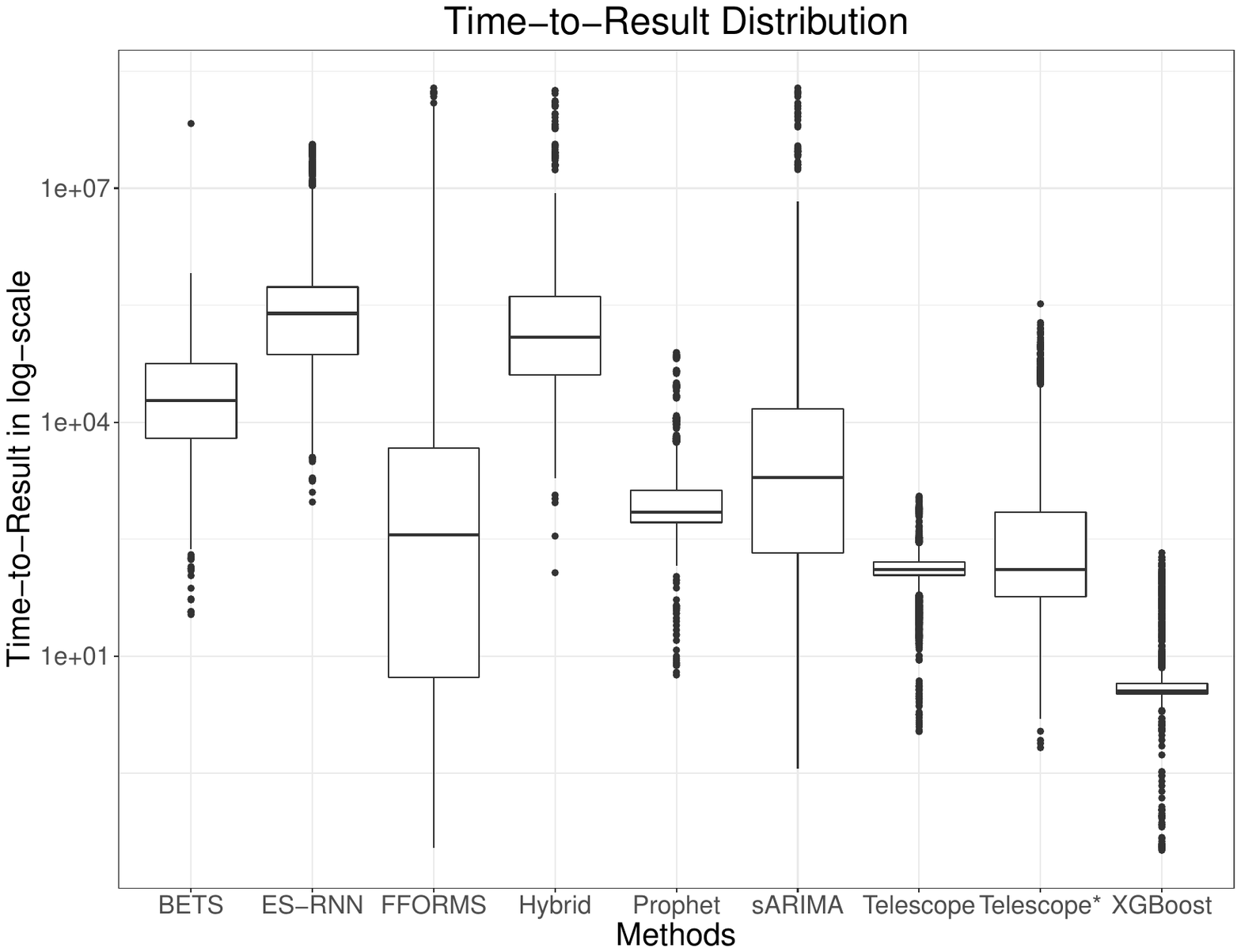}
	\caption{Time-to-result distribution.}
	\label{fig:timedis}
\end{figure}

Although the mean and standard deviation are useful statistical measures, we also examine the distributions of the forecast error and the time-to-result. Figure~\ref{fig:errordis} illustrates the forecast error distribution of all methods. Each distribution is depicted as a box plot, where the horizontal axis shows the different methods and the vertical axis the forecast error in log-scale. The methods FFORMS, sARIMA, Telescope, and Telescope$^*$ are showing almost the same distribution (i.e., quite short and similar interquartile ranges) with only a few outliers. Although Hybrid shows a similar distribution between the 25th and 75th quantiles, there are many outliers above the upper whisker. Another group with similar distribution comprises the methods BETS, Prophet, and XGBoost. The method with no outliers above the upper whisker but with the longest interquartile range is ES-RNN. However, the fairly similar error distributions are consistent with the ``No-Free-Lunch Theorem'', which states that there is no forecasting method that works best for all scenarios. Figure~\ref{fig:timedis} depicts the time-to-result distribution of all methods. Again, each distribution is illustrated as a box plot, and the vertical axis shows the time-to-result in log-scale. In contrast to the error distribution, the time-to-result distributions are completely different. XGBoost or Telescope exhibit a low variation in the time-to-result. In contrast,  FFORMS and sARIMA have a wide range of the time-to-result. Consequently, both methods may be impractical for time-critical scenarios. Telescope$^*$ has, in comparison to Telescope, also a huge variation regarding the time-to-result. Thus, Telescope is used in time-critical scenarios and Telescope$^*$ in non-time-critical scenarios.

\subsection{Detailed Examination}
\label{results:sec:telescope:sec:detailed}

Although our contribution Telescope can handle both seasonal and non-seasonal time series, it is intended for long and seasonal time series. To this end, we analyze the forecasting performance of all methods on seasonal (85\% of the data set) and non-seasonal (15\%) time series. Table~\ref{tab:season} lists the results for seasonal and non-seasonal time series. Each row shows a measure and each column a method. The best values (the lower, the better) are highlighted in bold. The lowest forecast error on average for seasonal time series is exhibited by Telescope$^*$ (19.77\%) followed by Telescope (20.33\%) and sARIMA (21.10\%). However, Telescope is about 6300 and Telescope$^*$ is about 260 times faster than sARIMA. In the case of non-seasonal time series, FFORMS (16.47\%) has the lowest forecast error followed by Telescope$^*$ (17.65\%) and Telescope (17.65\%). Note that both Telescope and Telescope$^*$ are using the same fallback for non-seasonal time series. Since the fallback, which comprises ARIMA, has a lower error than the sARIMA (17.85\%), we are able see the impact of the Preprocessing and Postprocessing phase of Telescope. In both cases, XGBoost exhibits the lowest time-to-result followed by Telescope.  

\begin{table*}[htb!]
	\centering
	\caption{Forecast error and time-to-result comparison on seasonal and non-seasonal time series.}
	\label{tab:season}
		\begin{tabular}{llrrrrrrrrr} 
			\toprule
			& \textbf{Measures}               & \textbf{BETS}   & \textbf{ES-RNN} & \textbf{FFORMS} & \textbf{Hybrid} & \textbf{Prophet} & \textbf{sARIMA} & \textbf{Telescope} & \textbf{Telescope$^*$} & \textbf{XGBoost} \\ \midrule
			\multirow{4}{*}{Seasonal}     & $\overline{e}$ [\%] & 26.79    & 52.30    & 21.94          & 29.40    & 38.12           & 21.10    & 20.33    & \textbf{19.77}  & 24.74            \\
			& $\sigma_{e}$ [\%]   & 33.24    & 71.18    & 39.11          & 92.16    & 2.49$\cdot10^2$ & 37.65    & 32.83    & \textbf{28.96}  & 36.73            \\
			& $\overline{t}_{N}$             & 7.14$\cdot10^4$ & 1.63$\cdot10^6$ & 6.11$\cdot10^5$       & 1.26$\cdot10^6$ & 2.22$\cdot10^3$        & 9.92$\cdot10^5$ & 1.56$\cdot10^2$ & 3.74$\cdot10^3$ & \textbf{6.90}    \\
			& $\sigma_{t_{N}}$               & 1.16$\cdot10^6$ & 4.92$\cdot10^6$ & 8.57$\cdot10^6$       & 8.73$\cdot10^6$ & 6.62$\cdot10^3$        & 9.81$\cdot10^6$ & 93.05    & 1.44$\cdot10^4$ & \textbf{15.71}    \\ \midrule
			\multirow{4}{*}{Non-Seasonal} & $\overline{e}$ [\%] & 18.00    & 21.74    & \textbf{16.47} & 18.98    & 20.53           & 17.85    & 17.65    & 17.65           & 18.59            \\
			& $\sigma_{e}$ [\%]   & 20.12    & 18.73    & 18.96          & 19.60    & 21.47           & 20.46    & 21.04    & 21.04           & \textbf{18.16}   \\
			& $\overline{t}_{N}$             & 2.53$\cdot10^3$ & 1.54$\cdot10^6$ & 1.74$\cdot10^2$       & 9.55$\cdot10^4$ & 9.26$\cdot10^2$        & 84.73    & 65.04    & 65.04           & \textbf{5.75}    \\
			& $\sigma_{t_{N}}$               & 4.42$\cdot10^3$ & 4.90$\cdot10^6$ & 8.20$\cdot10^2$       & 2.07$\cdot10^5$ & 1.07$\cdot10^3$        & 1.25$\cdot10^2$ & 95.05    & 95.05           & \textbf{14.83}    \\ \bottomrule 
		\end{tabular}
\end{table*}

\begin{figure}[htb!]
	\centering
	\includegraphics[width=0.75\linewidth]{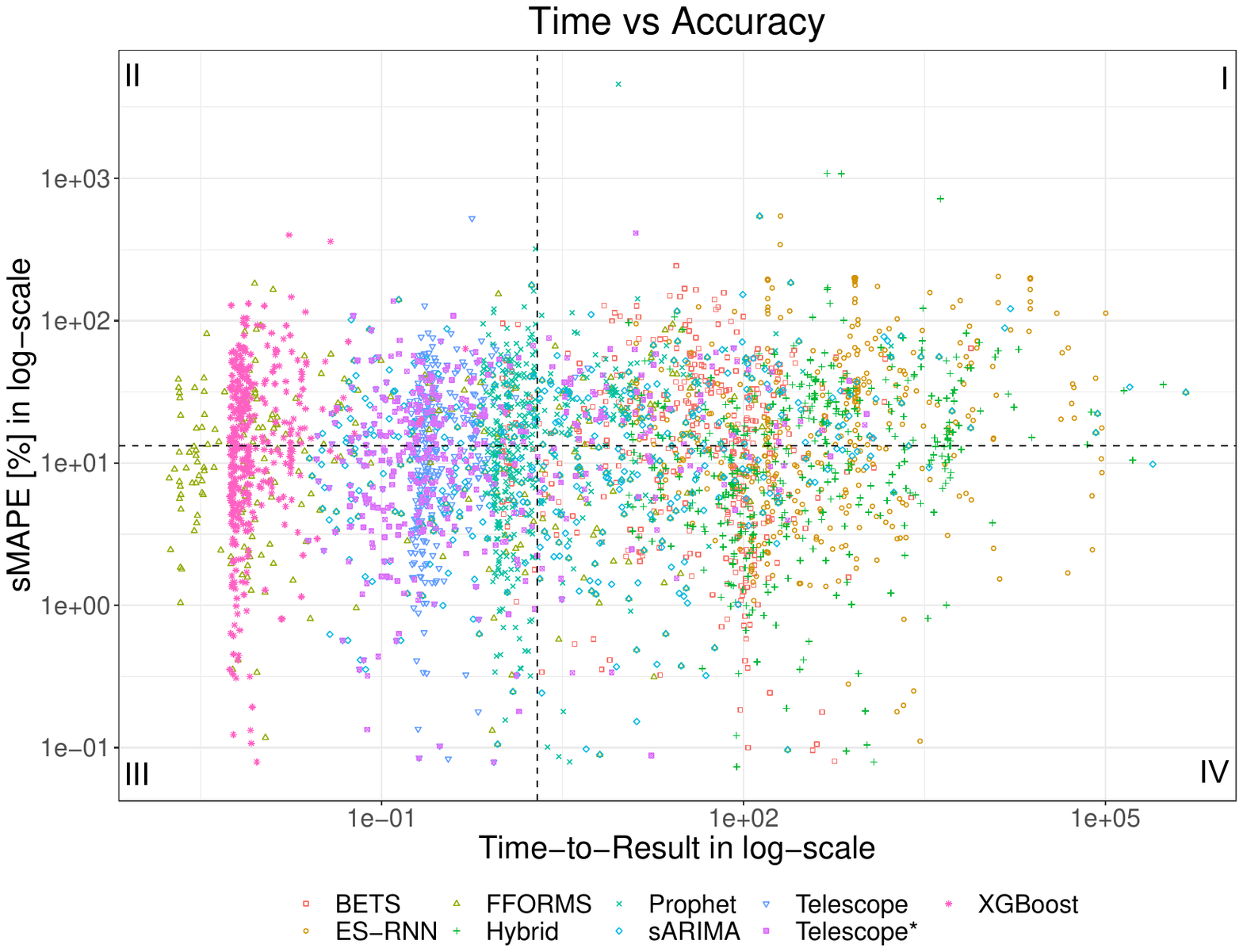}
	\caption{Forecast error vs. time-to-result for all methods.}
	\label{fig:accvstime}
\end{figure}

To investigate the trade-off between forecast error and time-to-result, we compare the methods in a 2-dimensional space spanned by the forecast error and time-to-result. We compute the median time-to-result~$\tilde{t}_{N}$ and median forecast error~$\tilde{e}$ over all methods. Based on these both values, we can sort the forecasts of each method for each time series $(t_{N},e)$ in one of the four quadrants: (i) $[\tilde{t}_{N}; \infty[\ \times\ [\tilde{e}; \infty[$, (ii) $[0; \tilde{t}_{N}[\ \times\ [\tilde{e}; \infty[$, (iii) $[0; \tilde{t}_{N}[\ \times\ [0; \tilde{e}[$, or (iv) $[\tilde{t}_{N}; \infty[\ \times\ [0; \tilde{e}[$. The best trade-off is achieved in the $3^{rd}$~quadrant as both the forecast error and time-to-result are lower than the median values. A semi-good performance is reflected by the $2^{nd}$ and $4^{th}$~quadrant as either the time-to-result or the forecast error is lower than the associated median value. The worst performance is achieved by forecasts in the $1^{st}$~quadrant. Here, both the forecast error and time-to-result are worse than the median values. Figure~\ref{fig:accvstime} depicts each forecast as a point in the 2-dimensional space. The vertical axis represents the forecast error and the horizontal axis the time-to-result. Both axes are in log-scale. The vertical dashed line represents $\tilde{t}_{N}$ and the horizontal axis $\tilde{e}$. Each forecasting method is depicted in an individual color and point shape. The methods XGBoost, Telescope, and Telescope$^*$ have compact clusters in the  $2^{nd}$ and $3^{rd}$~quadrant. In contrast, the methods BETS, ES-RNN, and Hybrid have far-reaching clusters in the $1^{st}$ and $4^{rd}$~quadrant. However, almost all methods have time series in each quadrant.

Table~\ref{tab:space} shows the distribution of the time series for each forecasting method in each quadrant. Each row represents a quadrant and each column a method. For instance, XGBoost has all forecasts equally distributed in the $2^{nd}$ and $3^{rd}$ quadrant. That is, all forecasts have a lower time-to-result than $\tilde{t}_{N}$. In contrast, ES-RNN and Hybrid have all of their forecasts in the $1^{st}$ and $4^{rd}$~quadrant. Consequently, all forecasts having a longer time-to-result than $\tilde{t}_{N}$. Telescope has 98\% of its forecast in the $2^{nd}$ and $3^{rd}$ quadrant. More precisely, 51\% of these forecasts are located in the $3^{rd}$ quadrant. In other words, Telescope has the highest number of forecasts exhibiting the best trade-off. The second most time series in the $3^{rd}$ quadrant has XGBoost (50\%) followed by Telescope$^*$ (42\%). 

\begin{table*}[htb!]
	\centering
	\caption{Distribution of time series in each quadrant for each forecasting method.}
		\begin{tabular}{llrrrrrrrrr} 
			\toprule
			&                                                & \textbf{BETS} & \textbf{ES-RNN} & \textbf{FFORMS} & \textbf{Hybrid} & \textbf{Prophet} & \textbf{sARIMA} & \textbf{Telescope} & \textbf{Telescope*} & \textbf{XGBoost} \\ \midrule
			QI:   & $[\tilde{t}_{N};\infty[ \times [\tilde{e};\infty[$ & 46\%           & 60\%             & 21\%             & 45\%             & 22\%              & 30\%             & 1\%        & 15\%                 & 0\%               \\
			QII:  & $[0;\tilde{t}_{N}[ \times [\tilde{e};\infty[$      & 4\%            & 0\%              & 26\%             & 0\%              & 33\%              & 17\%             & 47\%       & 33\%                 & 50\%              \\
			QIII: & $[0;\tilde{t}_{N}[ \times [0;\tilde{e}[$           & 5\%            & 0\%              & 34\%             & 0\%              & 36\%              & 22\%             & 51\%       & 42\%                 & 50\%              \\
			QIV:  & $[\tilde{t}_{N};\infty[ \times [0;\tilde{e}[$      & 45\%           & 40\%             & 19\%             & 55\%             & 10\%              & 31\%             & 1\%        & 9\%                 & 0\%               \\ \bottomrule
		\end{tabular}
	\label{tab:space}
\end{table*}

\subsection{Summary of the Results and Threats to Validity}
\label{results:sec:telescope:sec:threats}

Our experiments showed that both Telescope and Telescope$^*$ outperforms the state-of-the-art. More precisely, Telescope$^*$ achieves the best forecast accuracy followed by Telescope. Although our approach is intended for seasonal time series, it exhibits the second-best forecast accuracy on non-seasonal time series. Moreover, Telescope is, on average, up to 6000 times faster than the third most accurate method. In all experiments, Telescope$^*$ is more accurate than Telescope, but has, on average, a higher time-to-result as well as variation in the time-to-result. Also, we show that the chosen configuration of Telescope has the best trade-off between forecast accuracy and time-to-result. The third most accurate method, sARIMA, suffers from a high variation in the time-to-result. The winner of the M4-Competition, ES-RNN, is tailored for cross-learning to the M4-Competition data and therefore has the highest average forecast error. In summary, Telescope exhibits the best forecast accuracy coupled with a low and reliable time-to-result.

Although we applied a forecasting benchmark, the evaluation results may not be generalized to all time series from all areas. Besides the data set, we also try to have a sound set of recent hybrid forecasting methods based on different techniques. To this end, we also consider methods developed by Facebook and Uber. However, we use all methods with their default settings. Consequently, the observed results may differ if the forecasting methods are tuned to each time series. Moreover, our classification of the time series into long time series and time series with long periods may also affect the results. As Telescope achieves on the whole data set the best performance, the ranking inside the classes may only change. Lastly, we analyze whether the observed forecast accuracy as well as the measured time-to-result are statistically significant. To this end, we apply a non-parametric statistical test. More formally, we use the Friedman test~\cite{friedman1937use}, which ranks the forecasting methods separately for each time series and compare the average ranks of the methods. In case of a tie, average ranks are assigned. Thus, we formulate the following hypothesis:   
\begin{align}
H_{0,i}: \text{The methods perform equally} \nonumber
\end{align}
for the forecast error (i=1) and for the time-to-result (i=2). We conduct both hypotheses with a significance level of 1\%. The resulting p-values $p_1 < 2\cdot10^{-16}$ and $p_2 < 2\cdot10^{-16}$ indicate that both hypotheses can be rejected. Thus, the differences in the exhibited performance of the forecasting methods are statistically significant.

\section{Conclusion}
\label{sec:Conclusion}

This paper introduces Telescope, a novel machine learning-based forecasting approach that automatically retrieves relevant information from a given time series and splits it into parts, handling each of them separately (addressing RQ1).  More precisely, Telescope automatically extracts intrinsic time series features and then decomposes the time series into components, building a forecasting model for each of them (addressing RQ2). Each component is forecast by applying a different method and then the final forecast is assembled from the forecast components by employing a regression-based machine learning algorithm. For non-time-critical scenarios, we additionally provide an internal recommendation system that can be employed to automatically select the most appropriate machine learning algorithm for assembling the time series from its components (addressing RQ3).
	
In more than 1000 hours of experiments comparing Telescope against seven competing methods (including approaches from Uber and Facebook) using a forecasting benchmark, Telescope outperformed all methods, exhibiting the best forecast accuracy coupled with a low and reliable time-to-result. Compared to the competing methods that exhibited, on average, a forecast error (more precisely, the symmetric mean absolute error) of 29\%, Telescope exhibited an error of 20\% while being 2556~times faster. In particular, the methods from Uber and Facebook exhibited an error of 48\% and 36\%, and were 7334 and 19 times slower than Telescope, respectively.  When additionally applying the recommendation system, Telescope was able to reduce the forecast error even further down to 19\%. 
	
We see potential for extending Telescope to support multivariate time series and to detect structural changes in time series. The forecasting of multivariate time series has advantages and disadvantages. On the one hand, there is additional information that can be used to refine the prediction model. On the other hand, each extra piece of information has to be predictable to form the final forecast. If structural changes occur in time series, Telescope is prone to poor forecasts as it assumes, for example, that the seasonal pattern does not change over time. Besides the change in the seasonal pattern, further structural changes include level shifts or breakpoints in the trend. To mitigate this problem, structural changes have to be detected automatically. Based on the found changes, the values before the last change can be discarded if there is no regularity in the changes to consider only the time series's current structure for forecasting the time series.

\bibliographystyle{unsrt}  
\bibliography{bib}

\end{document}